\documentclass[11pt]{article}

\usepackage[preprint]{acl}

\usepackage{times}
\usepackage{latexsym}

\usepackage[T1]{fontenc}

\usepackage[utf8]{inputenc}

\usepackage{microtype}

\usepackage{inconsolata}

\usepackage{graphicx}
\usepackage{amsmath}
\usepackage{amssymb}
\usepackage{booktabs}
\usepackage{multirow}

\usepackage{float}

\usepackage{svg}
\usepackage{caption}
\usepackage{hyperref}

\usepackage{arydshln}
\usepackage{makecell}
\usepackage{array}
\usepackage{adjustbox}
\usepackage{subcaption}
\usepackage{tikz}
\usepackage[table]{xcolor}
\usepackage{colortbl}
\usepackage{pifont}
\usetikzlibrary{arrows.meta}

\usepackage{xcolor}      
\usepackage{tcolorbox}   
\usepackage{soul}        

\title{PoC: Performance-oriented Context Compression for Large Language Models via Performance Prediction}

\author{Runsong Zhao\textsuperscript{1,3} \quad
        Shilei Liu\textsuperscript{3} \quad
        Jiwei Tang\textsuperscript{2,3} \\
        \textbf{Langming Liu}\textsuperscript{3} \quad
        \textbf{Haibin Chen}\textsuperscript{3} \quad
        \textbf{Weidong Zhang\textsuperscript{3}} \quad
        \textbf{Yujin Yuan\textsuperscript{3}} \\
        \textbf{Tong Xiao\textsuperscript{1}\thanks{Corresponding authors.}} \quad
        \textbf{Jingbo Zhu\textsuperscript{1}} \quad
        \textbf{Wenbo Su\textsuperscript{3}} \quad
        \textbf{Bo Zheng\textsuperscript{3}\footnotemark[1]} \\
        \\
        \textsuperscript{1}Northeastern University, China \quad
        \textsuperscript{2}Tsinghua University \quad
        \textsuperscript{3}Future Living Lab of Alibaba \\
        \texttt{zhaors@mails.neu.edu.cn}
}

\begin{document}
\maketitle
\begin{abstract}

While context compression can mitigate the growing inference costs of Large Language Models (LLMs) by shortening contexts, existing methods that specify a target compression ratio or length suffer from \emph{unpredictable} performance degradation, hindering their reliable deployment. We introduce a paradigm shift to \textbf{P}erformance-\textbf{o}riented Context \textbf{C}ompression (\textbf{PoC}), where developers specify an acceptable performance floor instead of a compression ratio. PoC employs a lightweight performance predictor to automatically find the most aggressive compression ratio that satisfies this constraint before steering an off-the-shelf compressor. We design and compare two predictor variants: a simple \textit{context-agnostic} predictor and a more sophisticated \textit{context-aware} one that considers the input's inherent compressibility. On both question-answering and summarization benchmarks, the context-aware predictor consistently achieves lower performance prediction error than the context-agnostic predictor, while the resulting context-aware PoC attains a superior overall performance. Our work paves the way for a more reliable, efficient, and performance-aware deployment of context compression for LLMs.

\end{abstract}

\section{Introduction}

As large language models (LLMs) become increasingly prevalent in applications such as document question answering~\cite{pandya2021question} and summarization~\cite{zhang2025systematic}, their ability to process long-context information has become critical~\cite{liu-etal-2024-forgetting, zhao2026comet}. However, longer contexts come at a steep cost: dramatically increased inference latency and computational overhead. To address this, context compression~\cite{chang2024efficient, li-etal-2025-prompt,tang2025perceptioncompressortrainingfreeprompt,tang2025gmsa,zhang2025long,zhao-etal-2025-position,liu2026autoencodingfree,tang2026comicoarsetofinecontextcompression,lv2026datadistributionmattersdatacentric} has emerged as a widely adopted solution that can significantly accelerate the inference process.

\begin{figure*}
    \centering
    \includegraphics[width=0.85\linewidth]{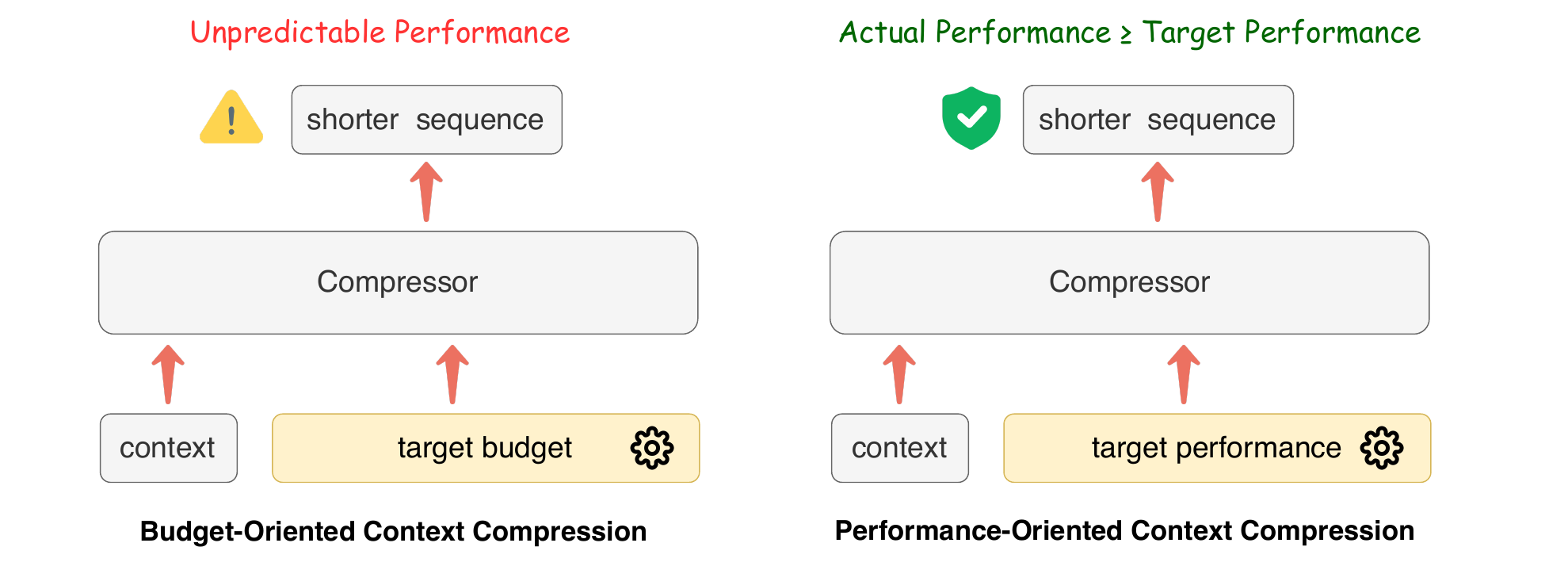}
    \caption{Overview of \textbf{Performance-oriented Context Compression (PoC)}. PoC is a novel context compression paradigm that moves away from budget-driven compression toward performance-driven compression. Unlike Budget-oriented Context Compression (BoC), PoC compresses context to the minimum length required to meet a given performance constraint. This approach mitigates the unpredictable performance degradation inherent in BoC, thus enhancing the reliability and applicability of context compression.}
    \label{fig:diag}
\end{figure*}

Context compression is a promising acceleration technique, but existing approaches are typically \textbf{Budget-oriented Context Compression (BoC)}, requiring a pre-specified compression ratio or length. The core limitation of this paradigm is that the relationship between compression ratio and task performance is complex and highly input-dependent, making performance degradation in BoC unpredictable. This unpredictability in BoC creates a dilemma at deployment time: \emph{a conservative ratio sacrifices efficiency gains on simple samples, while an aggressive ratio risks unacceptable performance loss on information-dense ones.} Consequently, uncontrollable performance degradation is the primary barrier preventing the widespread adoption of context compression techniques.

\begin{figure}[h]
    \centering
    \includegraphics[width=1\linewidth]{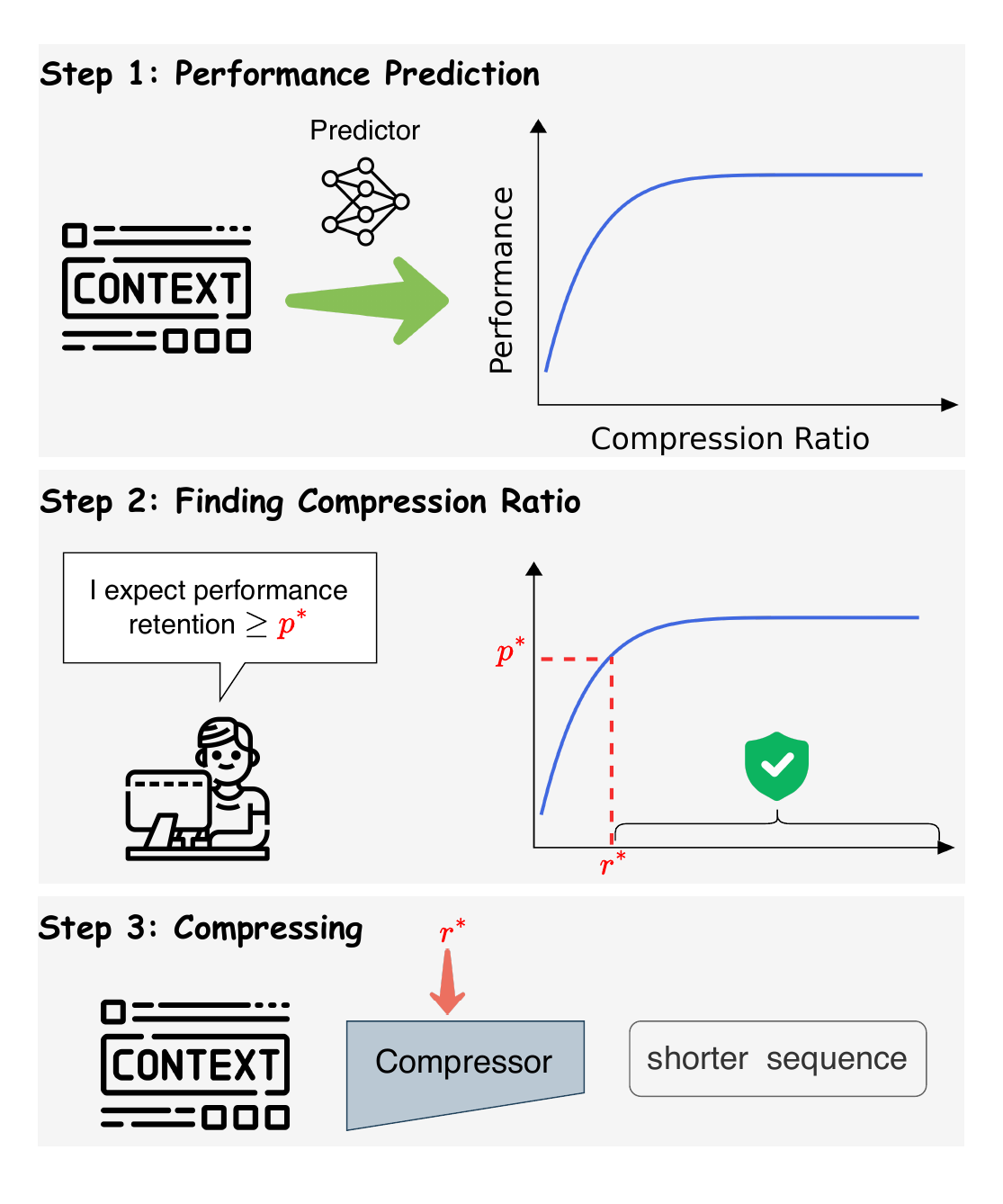}
    \caption{Overview of the PoC pipeline: (1) A performance predictor first estimates the performance-compression curve for the input context. (2) Given a target performance retention, PoC then searches for the minimum compression ratio that meets the constraint. (3) Finally, an off-the-shelf compressor compresses the context using the selected ratio.}
    \label{fig:method}
\end{figure}

To address this challenge, we propose \textbf{Performance-oriented Context Compression (PoC)}, which shifts control from the ``compression budget'' to ``task performance.'' As illustrated in Figure~\ref{fig:diag}, rather than setting a compression ratio\footnote{Let \(C\) and \(\tilde{C}\) denote the full and compressed contexts, respectively. We define the compression ratio as \(r = |\tilde{C}| / |C|\), where \(|\cdot|\) denotes token length. Smaller \(r\) implies more aggressive compression, and \(r = 1\) indicates no compression.} as in BoC, PoC compresses the context subject to a user-specified performance lower bound. To achieve this, we introduce a lightweight \textbf{performance predictor} (Figure~\ref{fig:method}) that estimates task performance of an off-the-shelf compressor at varying compression ratios, and selects the most aggressive ratio that satisfies the constraint.
We implement two predictor variants: a simple context-agnostic predictor that fits an average performance-compression curve\footnote{The x-axis of this curve represents the compression ratio, and the y-axis represents the performance retention.} on a calibration dataset, thereby serving as a natural extension of BoC, and a more sophisticated context-aware predictor that accounts for the intrinsic compressibility of each individual context. This context-aware variant transforms PoC into an adaptive compression framework, dynamically tailoring the compression ratio based on each input's unique properties.

A key challenge in evaluating adaptive compression methods is that they may operate at different compression ratios, so raw performance is therefore not directly comparable across methods.
To address this, we introduce the Performance at Ratio (P@R) metric to evaluate overall performance across a full range of compression ratios.
Across both question answering and summarization tasks, the context-aware predictor consistently yields lower performance prediction error and a better overall performance. Meanwhile, the context-aware predictor introduces only 1.53\,ms of latency, which is merely 5.6\% of the compressor's latency. This lightweight design makes it suitable for \textit{online} context compression scenarios, rather than being restricted to \textit{offline} RAG~\cite{lewis2020retrieval} settings. 

Our main contributions are summarized as follows:
\begin{itemize}
    \item We introduce a \textbf{performance-oriented context compression} framework that allows users to specify a target performance floor, making compression behavior more predictable and controllable.
    \item We propose the \textbf{P@R} metric to evaluate overall performance of adaptive compression frameworks, enabling fair comparison among methods operating at different compression ratios.
    \item We construct a training dataset for learning a \textit{context-aware} performance predictor.
    \item We empirically show that the \textit{context-aware} PoC framework achieves lower performance prediction error and a superior overall performance across compression ratios.
\end{itemize}

\section{Related Work}

To the best of our knowledge, our work is the first to introduce a context compression paradigm where the input is a target performance floor, rather than a compression budget. Since PoC is a meta-framework designed to steer any off-the-shelf \textbf{Context Compressor}, we first survey the landscape of these tools. We then position our context-aware variant relative to other approaches for \textbf{Adaptive Compression}, highlighting its unique advantages in versatility and predictability.

\paragraph{Context Compressors}
Context compression techniques can be broadly categorized into hard and soft compression. Hard compressors selectively remove less important tokens. Task-agnostic methods like LLMLingua~\cite{jiang2023llmlingua} and LLMLingua2~\cite{pan-etal-2024-llmlingua} assess token importance and discard non-essential tokens, while query-aware methods such as LongLLMLingua~\cite{jiang-etal-2024-longllmlingua}, Perception Compressor~\cite{tang2025perceptioncompressortrainingfreeprompt} and RECOMP~\cite{xu2024recomp} tailor compression to a specific query, at the cost of reduced generalizability. Soft compressors transform context into dense representations, either as compact embeddings~\cite{wingate-etal-2022-prompt,chevalier-etal-2023-adapting,ge2024incontext,tang2025gmsa} or as LLM KV-cache entries (e.g., GIST~\cite{mu2023learning}, Activation Beacon~\cite{zhang2025long}, 500xCompressor~\cite{li-etal-2025-500xcompressor}, EPL~\cite{zhao-etal-2025-position}, and SAC~\cite{liu2026autoencodingfree}). \emph{However, all soft methods require co-training with the LLM, rendering them incompatible with proprietary or black-box models. We therefore build PoC upon LLMLingua2 for its task-agnostic nature and broad compatibility.}

\paragraph{Adaptive Compression}

Several approaches explore adaptive compression via heuristics or reinforcement learning. QGC~\cite{cao-etal-2024-retaining} assigns higher compression ratios to more query-relevant documents; ACC-RAG~\cite{guo-etal-2025-enhancing} uses reinforcement learning to iteratively accumulate soft prompts; AdaComp~\cite{zhang2024adacomp} predicts how many documents to retrieve; and AttnComp~\cite{luo-etal-2025-attncomp} uses query-document attention to decide how many documents to retain. However, these methods are typically tied to specific compression mechanisms, require re-compression for each query, and often produce uncontrollable compression ratios, making direct comparison difficult.

Other methods~\cite{chen-etal-2025-dast,tang2026read,tang2026comicoarsetofinecontextcompression} introduce finer-grained adaptivity, but still assume a \emph{fixed} overall compression budget per sample. For instance, DAST~\cite{chen-etal-2025-dast} and COMI~\cite{tang2026comicoarsetofinecontextcompression} allocate soft token budgets according to the informativeness of different context segments, while keeping the sample-level compression ratio fixed. \emph{In contrast, our context-aware PoC is a more versatile and practical solution: it enhances existing off-the-shelf compressors, supports low-latency parallel prediction for online use, is query-agnostic, and enables controllable compression through a predictable performance floor.}

\begin{figure}
    \centering
    \includegraphics[width=1\linewidth]{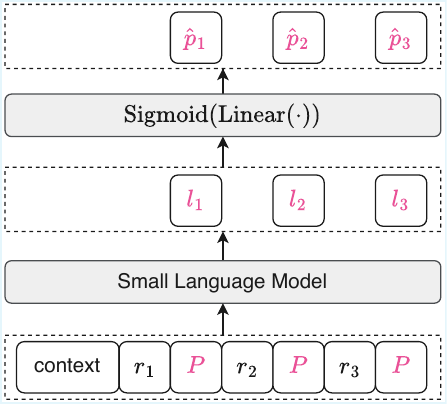}
    \caption{Context-aware predictor. It takes the context and multiple compression ratios as input to simultaneously predict multiple performance retentions. This parallel prediction approach enables the prediction of multiple performance retentions in a single forward pass, significantly accelerating the prediction speed.}
    \label{fig:aware_predictor}
\end{figure}

\section{Method}
\label{sec:method}

In this section, we first formalize the Performance-oriented Context Compression (PoC) framework in Section~\ref{sec:poc_framework}. We then detail the design and training of the performance predictor, a core component of our framework, in Section~\ref{sec:predictor}. Finally, Section~\ref{sec:search} describes an efficient search algorithm for determining the optimal compression ratio.

\subsection{PoC}
\label{sec:poc_framework}
As illustrated in Figure~\ref{fig:method}, the PoC pipeline operates in three main stages:

\paragraph{Performance Prediction}
Given an input context $C$, the primary step is to estimate its performance-compression curve. A performance predictor, the design of which is detailed in Section~\ref{sec:predictor}, takes the context and a set of $N$ candidate compression ratios $\{r_i\}_{i=1}^N$ as input to output the corresponding predicted performance retentions $\{\hat{p}_i\}_{i=1}^N$:
\begin{equation}
    \{\hat{p}_i\}_{i=1}^N = \text{Predictor}(C, \{r_i\}_{i=1}^N) \, .
\end{equation}
For a simpler \textit{context-agnostic} variant, the prediction is independent of the context $C$ and depends only on the ratios.

\paragraph{Finding Compression Ratio}
With the predicted performance curve, PoC searches for the optimal compression ratio $r^*$. This search, performed using the two-stage algorithm described in Section~\ref{sec:search}, aims to find the smallest ratio (i.e., the most aggressive compression) that satisfies a user-specified performance floor $p^*$. This can be formulated as:
\begin{equation}
    r^* = \arg\min_{r_i} \{ r_i \mid \hat{p}_i \ge p^* \} \, .
\end{equation}

\paragraph{Context Compression}
Finally, an off-the-shelf compressor is used to shorten the original context $C$ to the selected ratio $r^*$\footnote{Long contexts are segmented into 512-token chunks, with prediction and compression applied chunk-wise.}. The resulting compressed context $\tilde{C}$ is then provided to the LLM, along with the original instruction $I$, to generate the final response $Y$:
\begin{align}
    \tilde{C} &= \text{Compressor}(C, r^*) \, , \\
    Y &= \text{LLM}(\tilde{C}, I) \, .
\end{align}
This paradigm shifts the control from a fixed compression budget to a predictable performance outcome, enhancing the reliability of the compression process.

\subsection{Performance Predictor}
\label{sec:predictor}

The effectiveness of PoC hinges on an accurate performance predictor. Here, we describe the process of training such a predictor, from data collection to model architecture.

\paragraph{Performance Retention as Prediction Target}
Directly predicting raw task scores (e.g., F1, ROUGE~\cite{lin-2004-rouge}) is challenging due to their varying scales and distributions across different tasks. To create a unified and generalizable prediction target, we define \textbf{performance retention} $p_i$ as the normalized score at compression ratio $r_i$:
\begin{equation}
p_i = \min\left(\frac{M_{r_i}}{M_1}, 1\right) \, ,
\label{eq:performance_retention}
\end{equation}
where $M_{r_i}$ is the task score with compression ratio $r_i$, and $M_1$ is the baseline score without compression ($r_i=1$). This metric, ranging from 0 to 1, intuitively represents the fraction of performance preserved after compression.

\paragraph{Data Collection}
To train the predictor, we construct a dataset of $(C, \{r_i\}_{i=1}^N, \{p_i\}_{i=1}^N)$ tuples. We begin with a source dataset $\mathcal{D}$, where each data point is a tuple $(C, I, A)$ representing the context, instruction, and ground-truth answer, respectively. For each such sample, we generate a training instance for our predictor by following these steps:
\begin{enumerate}
    \item Sample a set of compression ratios $\{r_i\}_{i=1}^N$ from the interval $[0, 1]$.
    \item For each $r_i$, compress the context $C$ using an off-the-shelf compressor to obtain the compressed context $\tilde{C}_{i}$.
    \item Generate a response $\hat{A}_{i}$ from the reader LLM using the compressed context and the original instruction: $\hat{A}_{i} = \text{LLM}(\tilde{C}_{i}, I)$.
    \item Compute the raw score $M_{r_i}$ by evaluating the generated response $\hat{A}_{i}$ against the ground-truth answer $A$.
    \item Calculate the performance retention $p_i$ using Equation~\ref{eq:performance_retention}.
\end{enumerate}
This process yields the final training tuples required for our predictor. The specific datasets $\mathcal{D}$ and models used are detailed in our experimental setup in Section~\ref{sec:experiments}. The datasets we collected can be accessed from the submission materials.

\paragraph{Context-Agnostic Predictor}
As a simple baseline, we design a context-agnostic predictor. This predictor models the \textit{average} performance-compression trend over an entire calibration dataset $\mathcal{D}$. We first compute the mean performance retention at $N$ discrete compression ratios. Then, we use cubic spline interpolation to fit a continuous curve through these points. While this approach can provide a general performance estimate for any ratio, it ignores the unique compressibility of individual contexts.

\paragraph{Context-Aware Predictor}
To capture sample-specific characteristics, we propose a more sophisticated context-aware predictor, shown in Figure~\ref{fig:aware_predictor}. This model is based on a lightweight small language model and is designed for efficient, parallel prediction. For a given context $C$ and a set of candidate ratios $\{r_i\}_{i=1}^N$, we construct a single input sequence $X$:
\begin{equation}
X = C \oplus r_1 \oplus \texttt{[P]} \oplus \dots \oplus r_N \oplus \texttt{[P]} \, ,
\end{equation}
where $\oplus$ is sequence concatenation, $r_i$ is the text form of the ratio, and $\texttt{[P]}$ is a special token. The final hidden states $\{l_i\}_{i=1}^N$ corresponding to the $\texttt{[P]}$ tokens are extracted and passed through a linear layer followed by a sigmoid function to predict the performance retentions:
\begin{equation}
\hat{p}_i = \text{Sigmoid}(\text{Linear}(l_i)) \, .
\end{equation}
This parallel architecture allows for the simultaneous prediction of multiple points on the curve in a single forward pass, minimizing inference overhead. The model is trained with Mean Squared Error (MSE) loss.

\subsection{Two-Stage Search Algorithm}
\label{sec:search}
To efficiently find the optimal compression ratio $r^*$ with high precision, we employ a two-stage search algorithm that balances accuracy and computational cost.

\paragraph{Stage 1: Coarse-grained Search}
In the first stage, we perform a broad search across the entire $[0, 1]$ interval. We uniformly sample 18 candidate ratios and use our context-aware predictor to estimate their corresponding performance retentions in a single forward pass. From these predictions, we identify a narrow sub-interval that brackets the optimal solution $r^*$. This reduces the search space with an initial precision of approximately $1/19 \approx 0.053$.

\paragraph{Stage 2: Fine-grained Search}
In the second stage, we conduct a focused search within the sub-interval identified previously. We again sample 18 ratios, this time uniformly from the narrow sub-interval, and run a second prediction pass. The best ratio from this fine-grained search is selected as the final $r^*$. This two-stage process refines the search precision to approximately $1/19^2 \approx 0.0028$, allowing us to pinpoint the optimal compression ratio with minimal overhead.

\section{Experiments}
\label{sec:experiments}

\subsection{Datasets and Tasks}
Our experiments are conducted on a diverse suite of seven benchmarks spanning question answering (QA) and summarization tasks, featuring contexts with lengths up to 32,768 tokens.

For Question Answering, we use five standard datasets: SearchQA~\cite{dunn2017searchqa}, TriviaQA~\cite{joshi-etal-2017-triviaqa}, Natural Questions~\cite{kwiatkowski-etal-2019-natural}, HotpotQA~\cite{yang-etal-2018-hotpotqa}, and SQuAD~\cite{rajpurkar-etal-2016-squad}. For these tasks, we evaluate performance using the F1 score, which measures the harmonic mean of precision and recall between the predicted and ground-truth answers. Compared to Exact Match (EM), F1 yields a softer and more informative training signal for our performance predictor, as it captures partial overlaps between predictions and references rather than treating correctness as a binary outcome.

For Summarization, we use the GovReport~\cite{huang-etal-2021-efficient} and SummScreenFD~\cite{chen-etal-2022-summscreen} datasets. Performance is measured by the geometric mean of ROUGE-1, ROUGE-2, and ROUGE-L scores~\cite{lin-2004-rouge}, providing a balanced assessment of unigram overlap, bigram overlap, and the longest common subsequence.

\subsection{Implementation Details}
Our experimental setup is comprised of three core components.

\paragraph{Compressor} We employ LLMLingua2~\cite{pan-etal-2024-llmlingua}, a widely adopted task-agnostic hard compressor. Given a target compression ratio and context, it selectively retains the most crucial tokens from the original context.

\paragraph{Reader LLM} We use Llama-3.1-8B-Instruct as the reader model~\cite{grattafiori2024llama}. It takes the compressed context and the original instruction (e.g., a question or a summarization prompt) as input and generates the final response.

\paragraph{Performance Predictor} Our context-aware predictor is based on Llama-3.2-1B-Instruct. To create an efficient and lightweight predictor, we prune the model significantly, retaining only its token embedding layer and the initial two Transformer layers.

\paragraph{Predictor Training} We train the context-aware predictor using the AdamW optimizer~\cite{loshchilov2018decoupled} with a batch size of 256 and a learning rate of $2 \times 10^{-5}$. We apply a weight decay of 0.01 and use a linear learning rate scheduler with a warmup phase over the first 2\% of training steps. The entire training process leverages bfloat16 mixed-precision. From our collected data (as described in Section~\ref{sec:predictor}), we allocate 98\% for training and reserve the remaining 2\% for validation.

\begin{table*}[!t]
\centering
\small
\resizebox{\linewidth}{!}{
\begin{tabular}{l ccc ccc ccc ccc ccc}
\toprule
\multirow{2}{*}{\textbf{Method}} & \multicolumn{3}{c}{\textbf{SearchQA}} & \multicolumn{3}{c}{\textbf{TriviaQA}} & \multicolumn{3}{c}{\textbf{Natural Questions}} & \multicolumn{3}{c}{\textbf{HotpotQA}} & \multicolumn{3}{c}{\textbf{SQuAD}} \\
\cmidrule(lr){2-4} \cmidrule(lr){5-7} \cmidrule(lr){8-10} \cmidrule(lr){11-13} \cmidrule(lr){14-16}
 & PPE$\downarrow$ & F1@R$\uparrow$ & EM@R$\uparrow$ & PPE$\downarrow$ & F1@R$\uparrow$ & EM@R$\uparrow$ & PPE$\downarrow$ & F1@R$\uparrow$ & EM@R$\uparrow$ & PPE$\downarrow$ & F1@R$\uparrow$ & EM@R$\uparrow$ & PPE$\downarrow$ & F1@R$\uparrow$ & EM@R$\uparrow$ \\
\midrule
context-agnostic & 0.156 & 62.3 & 47.4 & 0.122 & 49.2 & 32.0 & 0.168 & 61.9 & 42.0 & 0.162 & 66.3 & 47.0 & 0.148 & 62.8 & 41.8 \\
context-aware & \textbf{0.136} (\textcolor{teal}{↓12.8\%}) & \textbf{64.7} & \textbf{50.4} & \textbf{0.113} (\textcolor{teal}{↓7.4\%}) & \textbf{67.5} & \textbf{53.1} & \textbf{0.156} (\textcolor{teal}{↓7.1\%}) & \textbf{62.2} & \textbf{42.3} & \textbf{0.159} (\textcolor{teal}{↓1.9\%}) & \textbf{67.0} & \textbf{47.5} & \textbf{0.144} (\textcolor{teal}{↓2.7\%}) & \textbf{63.0} & \textbf{42.4} \\
\bottomrule
\end{tabular}
}
\caption{Performance on Question Answering benchmarks. We report the Performance Prediction Error (PPE, lower is better) and Performance at Ratio (P@R) for F1 and EM scores (higher is better). The percentage in parentheses indicates the relative reduction in PPE compared to the context-agnostic baseline.}
\label{tab:qa_results_updated}
\end{table*}

\begin{table*}[!t]
\centering
\small
\begin{tabular}{l cccc cccc}
\toprule
\multirow{2}{*}{\textbf{Method}} & \multicolumn{4}{c}{\textbf{GovReport}} & \multicolumn{4}{c}{\textbf{SummScreenFD}} \\
\cmidrule(lr){2-5} \cmidrule(lr){6-9}
 & PPE$\downarrow$ & R1@R$\uparrow$ & R2@R$\uparrow$ & RL@R$\uparrow$ & PPE$\downarrow$ & R1@R$\uparrow$ & R2@R$\uparrow$ & RL@R$\uparrow$ \\
\midrule
context-agnostic & 0.023 & 34.0 & 13.2 & 18.1 & 0.059 & 29.5 & 5.8 & 16.2 \\
context-aware & \textbf{0.018} (\textcolor{teal}{↓21.7\%}) & 34.0 & 13.2 & 18.1 & \textbf{0.039} (\textcolor{teal}{↓33.9\%}) & \textbf{30.3} & 5.8 & \textbf{16.5} \\
\bottomrule
\end{tabular}
\caption{Performance on Summarization benchmarks. We report the Performance Prediction Error (PPE) and Performance at Ratio (P@R) for ROUGE-1/2/L scores. The percentage in parentheses shows the relative reduction in PPE for our context-aware model compared to the baseline.}
\label{tab:summ_results_updated}
\end{table*}

\subsection{Evaluation Metrics}
\label{sec:metric}
We assess the effectiveness of our framework using two primary metrics.

\paragraph{Performance Prediction Error (PPE)}
To measure the accuracy of our performance predictor, we calculate the Mean Squared Error (MSE) between the predicted performance retentions ($\hat{p}_i$) and the ground-truth retentions ($p_i$) on the validation set. A lower MSE indicates a more accurate predictor, which is crucial for the PoC framework to reliably meet the specified performance floor. The PPE is defined as:
\begin{equation}
\text{PPE} = \frac{1}{N}\sum_{i=1}^{N} (p_i - \hat{p}_i)^2 \, .
\end{equation}

\paragraph{Performance at Ratio (P@R)}
To evaluate overall performance across compression ratios, we introduce the Performance at Ratio (P@R) metric. We derive this metric through the following procedure: first, we sample a range of target performance floors. For each floor, we run our PoC framework over the entire test set and compute the resulting average compression ratio and average achieved task performance. This process yields a set of (average ratio, average performance) data points. We then use cubic spline interpolation to fit a continuous performance-compression curve, $p = f(r)$. Finally, we define P@R as the Area Under this Curve (AUC), which represents the expected performance if the compression ratio $R$ were drawn from a uniform distribution over $[0, 1]$. The P@R is formally expressed as:
\begin{equation}
\text{P@R} = \mathbb{E}_{R \sim U(0,1)}[P] = \int_{0}^{1} f(r) \, dr \, .
\label{eq:par}
\end{equation}
A higher P@R value indicates superior average performance across the entire spectrum of possible compression ratios.

\subsection{Results}

Table~\ref{tab:qa_results_updated} presents the results on the five question answering benchmarks. Our context-aware PoC framework consistently outperforms the context-agnostic baseline across all datasets. A key advantage is the significantly improved prediction accuracy, as evidenced by the lower Performance Prediction Error (PPE). For instance, the context-aware predictor reduces PPE by 12.8\% on SearchQA and 7.4\% on TriviaQA. This stems from its ability to account for the intrinsic, sample-specific relationship between compression ratio and performance; an information-dense context requires a less aggressive compression ratio to maintain performance compared to a sparse one, a nuance the context-agnostic baseline cannot capture. This superior prediction accuracy translates directly to a better overall performance, reflected in higher Performance at Ratio (P@R) scores. On SearchQA, the context-aware approach improves F1@R from 62.3 to 64.7 and EM@R from 47.4 to 50.4. The improvement is particularly dramatic on TriviaQA, where F1@R soars from 49.2 to 67.5, and EM@R from 32.0 to 53.1. We attribute this substantial gain to the unique characteristics of the TriviaQA dataset and will delve into this further in our analysis in Section~\ref{sec:analysis_tradeoff}. The core reason for the improved performance is that by accurately predicting performance, our method can adaptively apply the most aggressive compression possible for each sample without violating the performance floor, thus optimizing overall efficiency.

The results for the summarization tasks, shown in Table~\ref{tab:summ_results_updated}, further underscore the benefits of context-awareness. The PPE reduction is even more pronounced here, with decreases of 21.7\% on GovReport and 33.9\% on SummScreenFD. Notably, the absolute PPE values for summarization (e.g., <0.06) are substantially lower than for the QA tasks (>0.12), suggesting that the performance-compression curve in summarization is more consistent and easier to model. In terms of the overall performance, the context-aware method shows modest gains on SummScreenFD and matches the baseline on GovReport. This can be attributed to the nature of summarization, which typically requires a holistic understanding of the entire context. Unlike QA where key information may be localized, most parts of the context are often relevant for generating a good summary. Consequently, the potential for aggressive, adaptive compression varies less across samples, leading to smaller, though still positive, improvements in the P@R metric compared to QA.

\section{Analysis}



\begin{figure*}[t]
\centering
\captionsetup[sub]{font=small, labelfont=small}

 \begin{minipage}{0.25\textwidth}
        \centering
        \subcaptionbox{SearchQA\label{fig:SearchQA}}{
         \includegraphics[width=\linewidth]{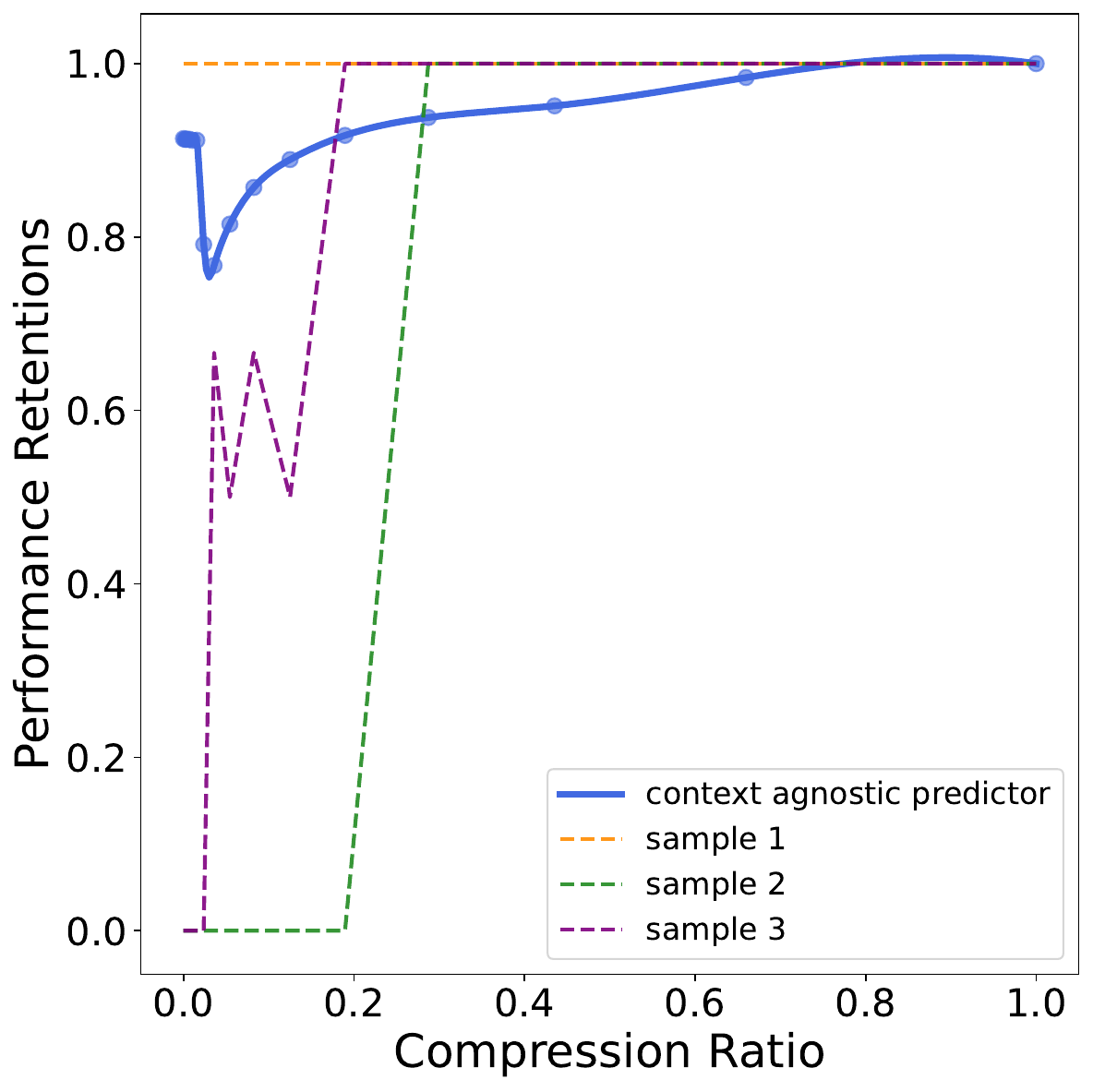}
        }
 \end{minipage}%
 \begin{minipage}{0.25\textwidth}
        \centering
        \subcaptionbox{TriviaQA\label{fig:TriviaQA}}{
         \includegraphics[width=\linewidth]{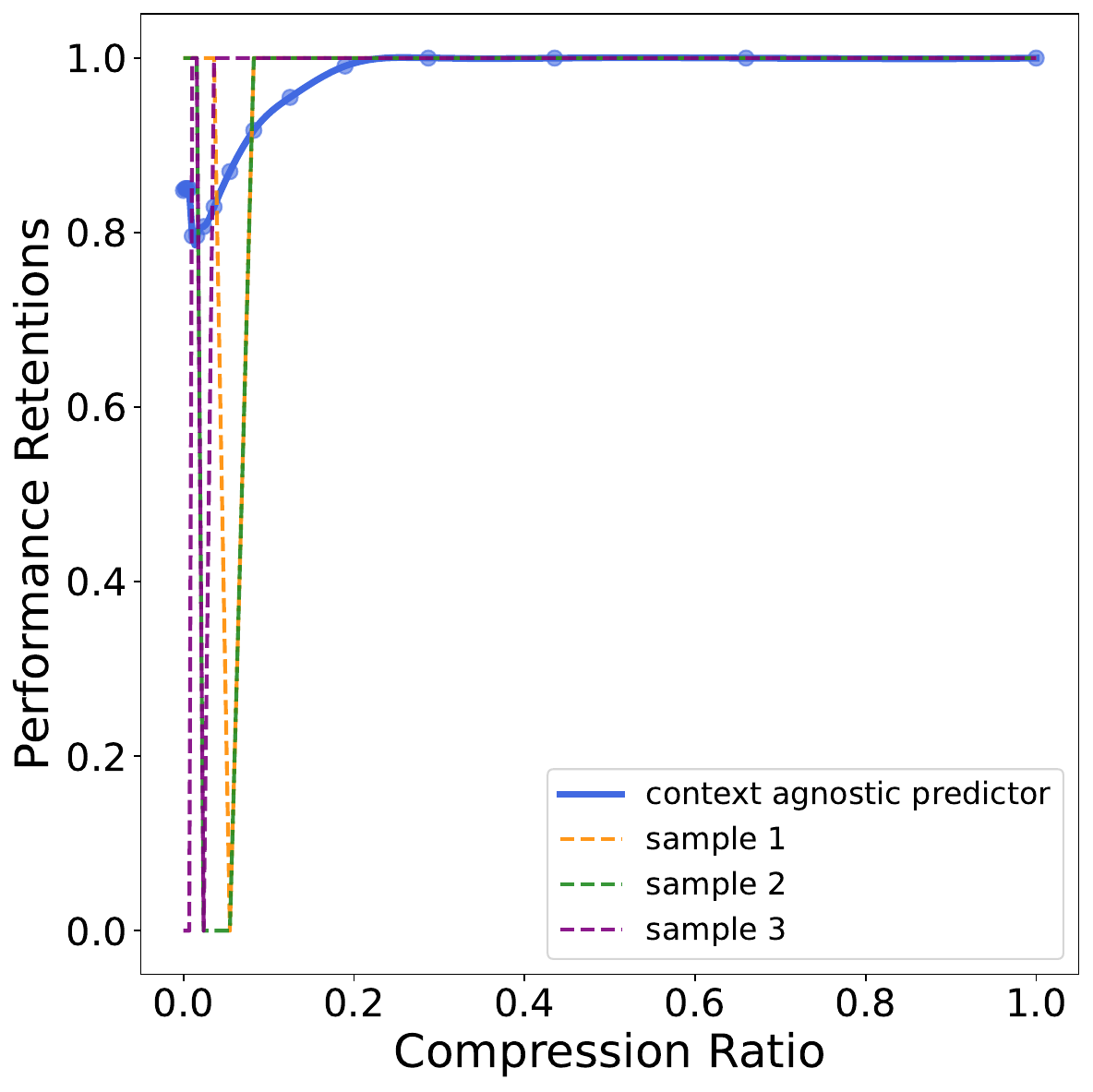}
        }
 \end{minipage}%
 \begin{minipage}{0.25\textwidth}
    \centering
    \subcaptionbox{Natural Questions\label{fig:NQ}}{
    \includegraphics[width=\linewidth]{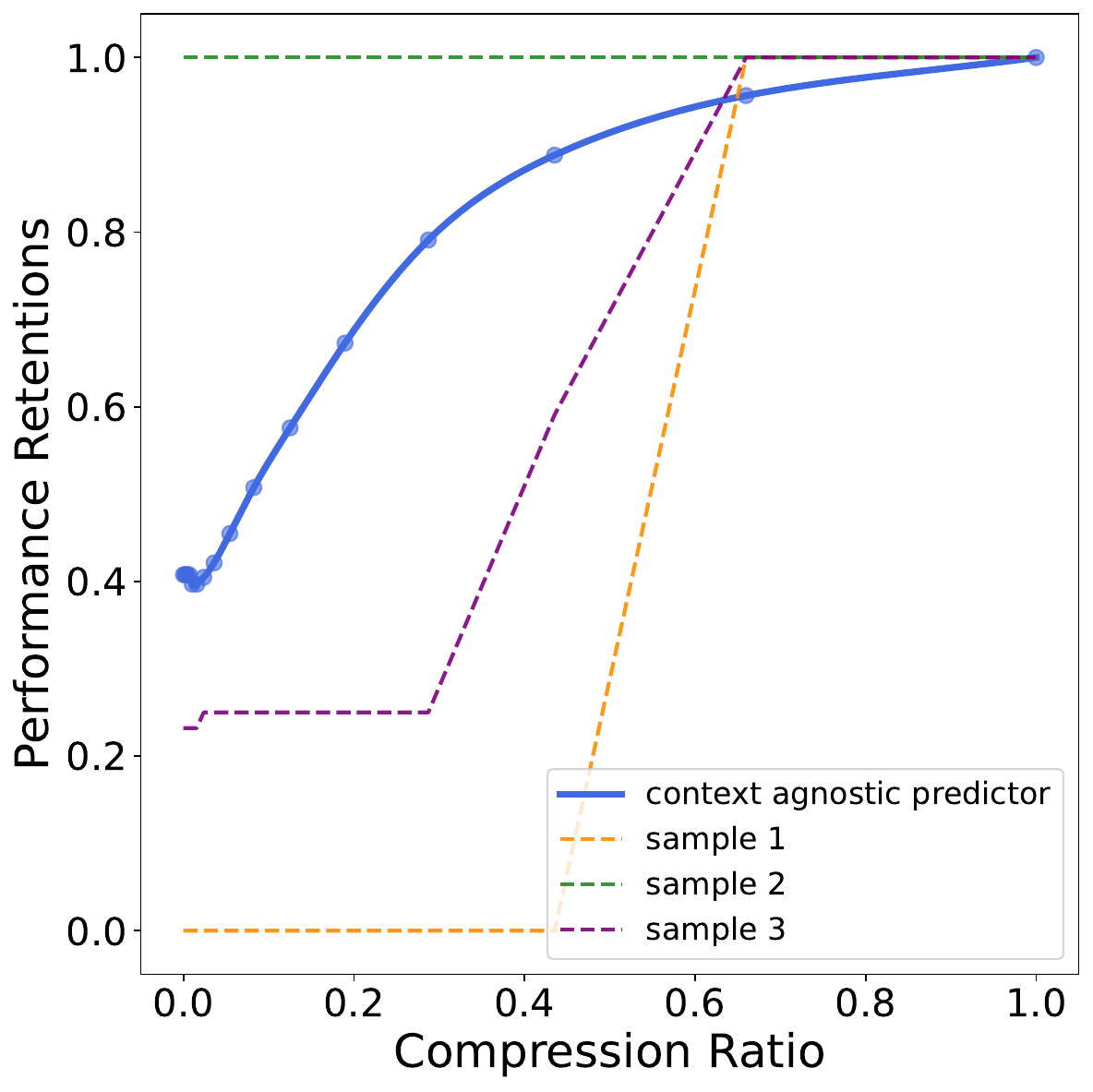}
    }
\end{minipage}%
\begin{minipage}{0.25\textwidth}
    \centering
    \subcaptionbox{HotpotQA\label{fig:HotpotQA}}{
    \includegraphics[width=\linewidth]{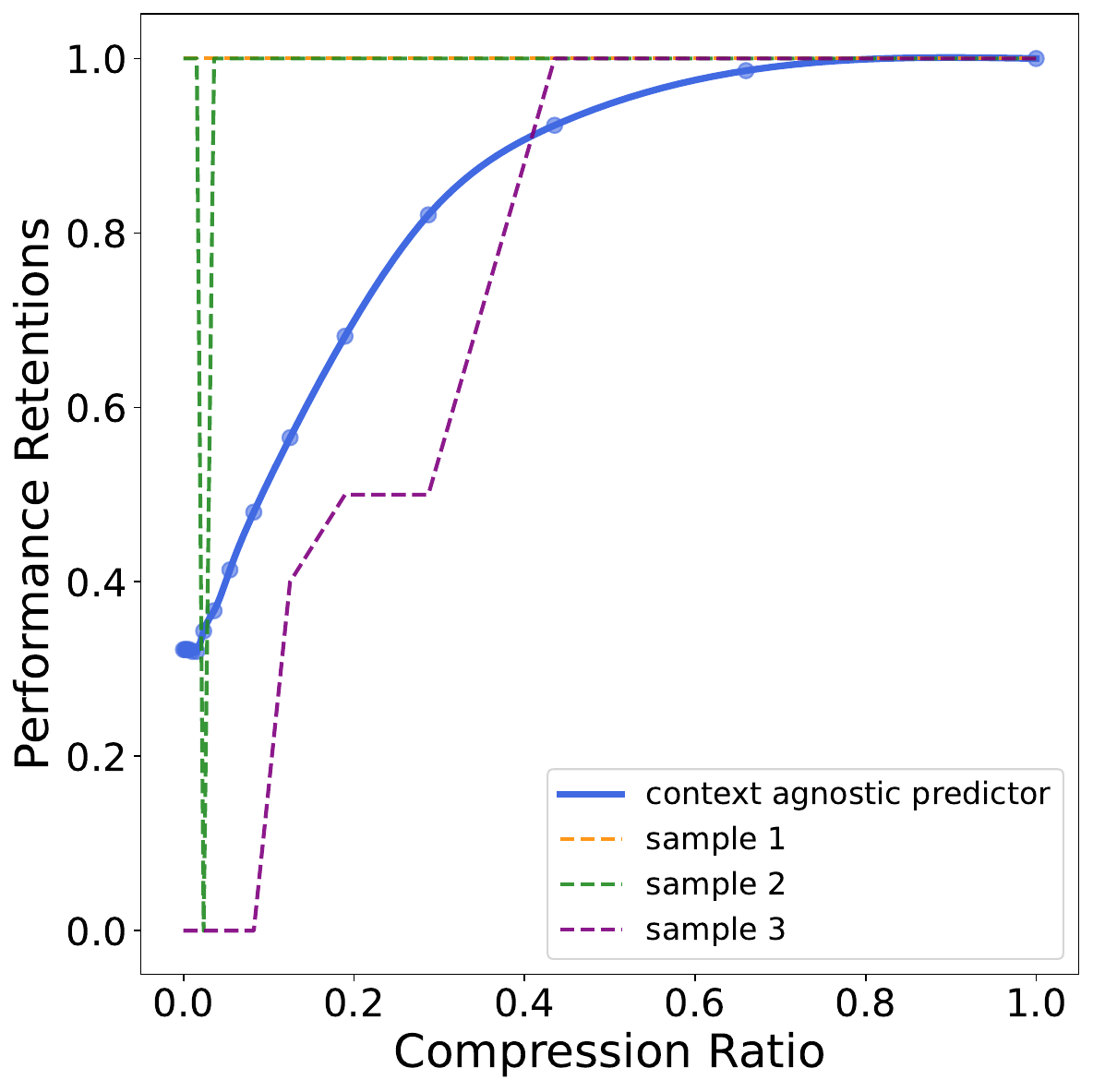}
    }
\end{minipage}

\vspace{5mm}

 \begin{minipage}{0.25\textwidth}
        \centering
        \subcaptionbox{SQuAD\label{fig:SQuAD}}{
         \includegraphics[width=\linewidth]{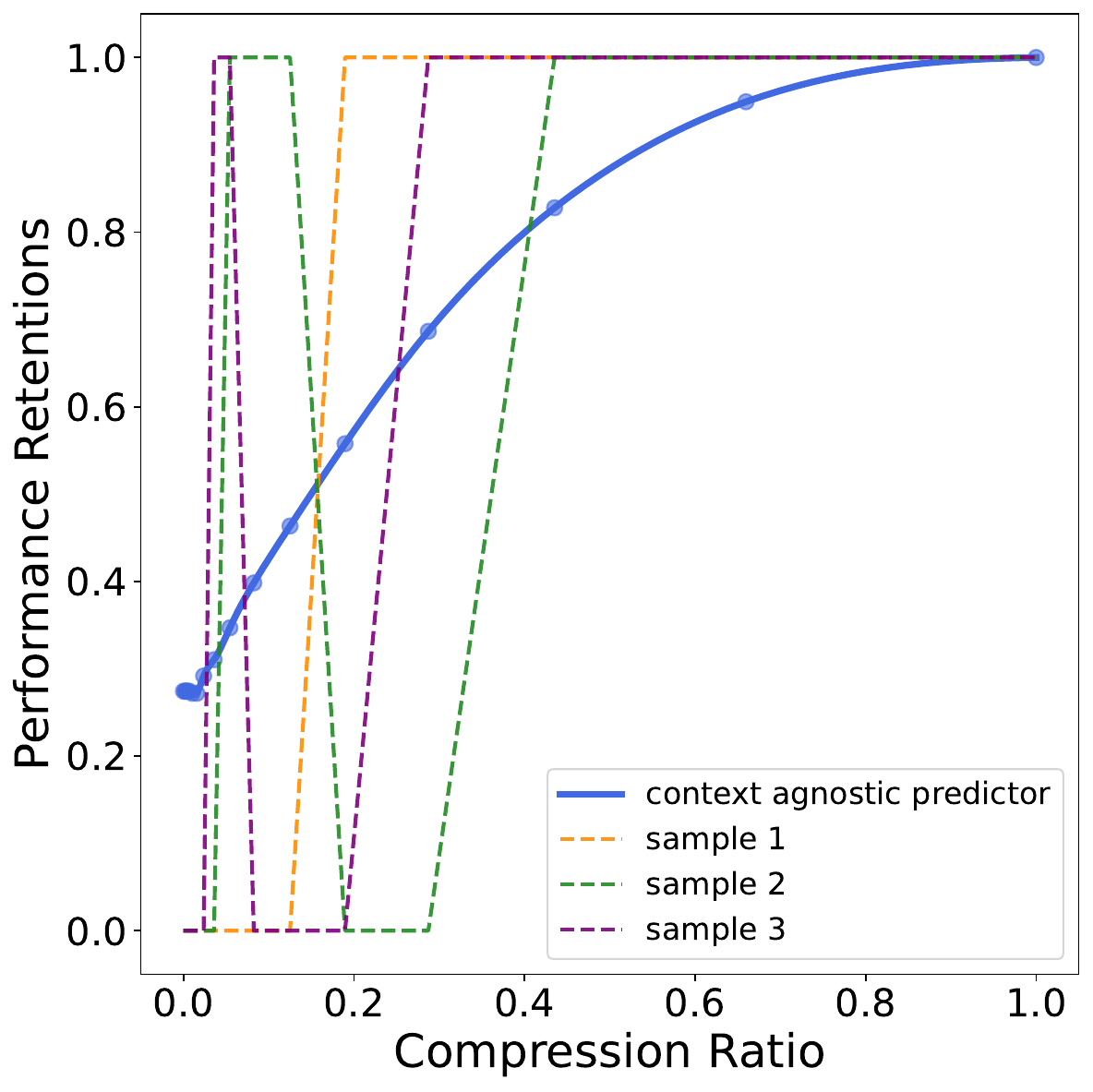}
        }
 \end{minipage}%
 \begin{minipage}{0.25\textwidth}
        \centering
        \subcaptionbox{GovReport\label{fig:gov_report}}{
         \includegraphics[width=\linewidth]{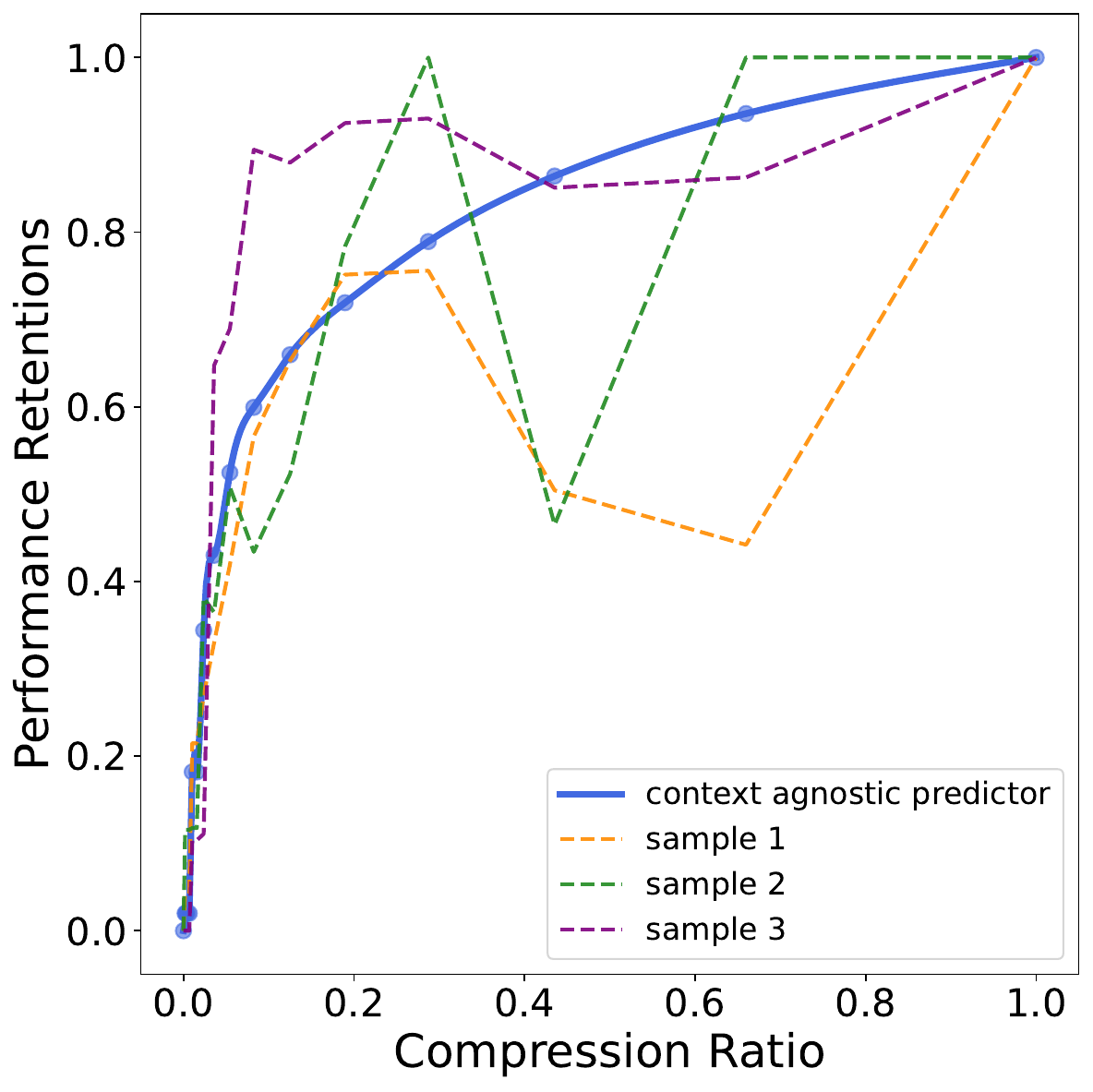}
        }
 \end{minipage}%
 \begin{minipage}{0.25\textwidth}
    \centering
    \subcaptionbox{SummScreenFD\label{fig:summ_screen_fd}}{
    \includegraphics[width=\linewidth]{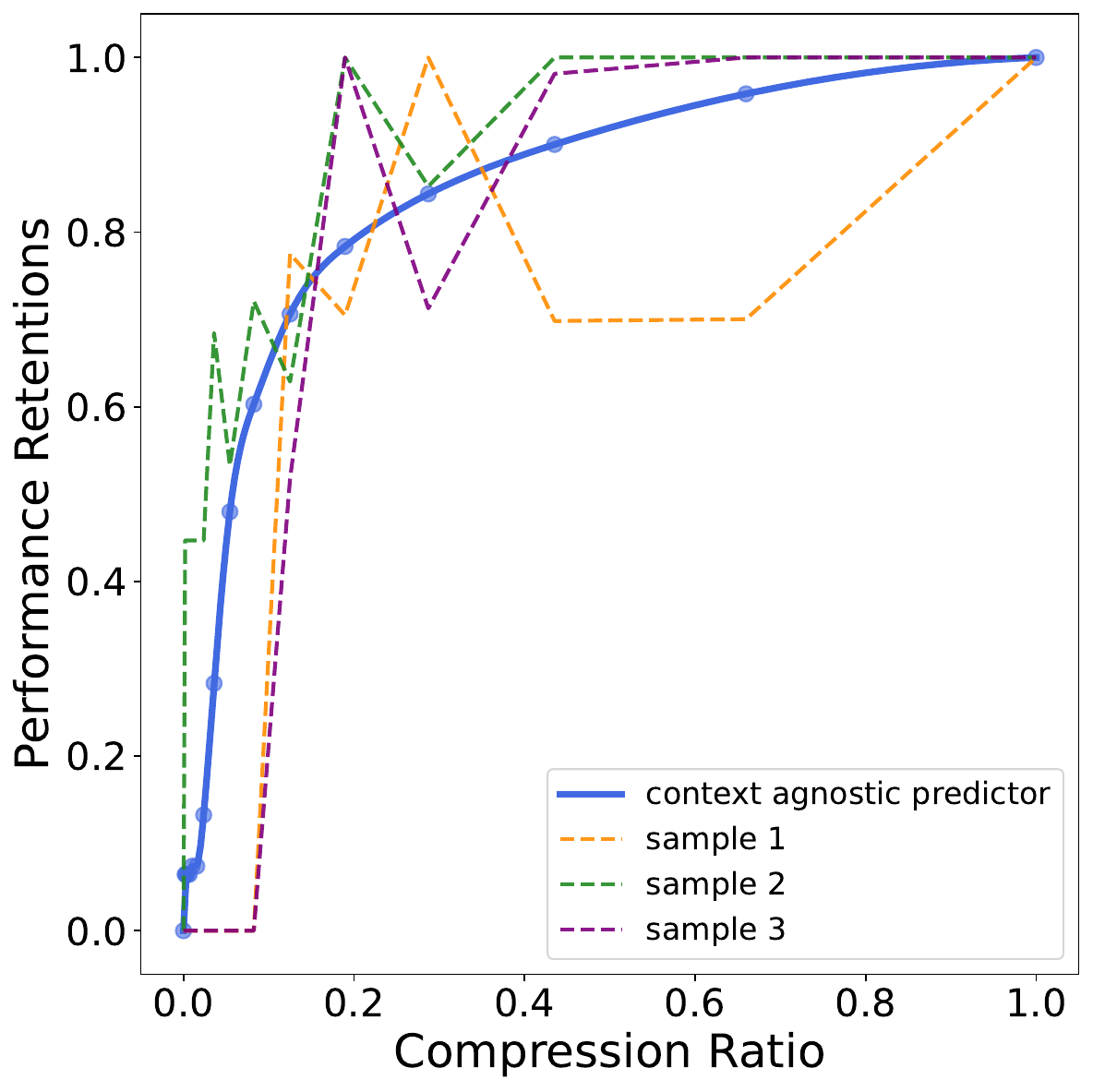}
    }
\end{minipage}%
\begin{minipage}{0.25\textwidth}
    \centering
    \subcaptionbox{TriviaQA\label{fig:TriviaQA_f1_s}}{
    \includegraphics[width=\linewidth]{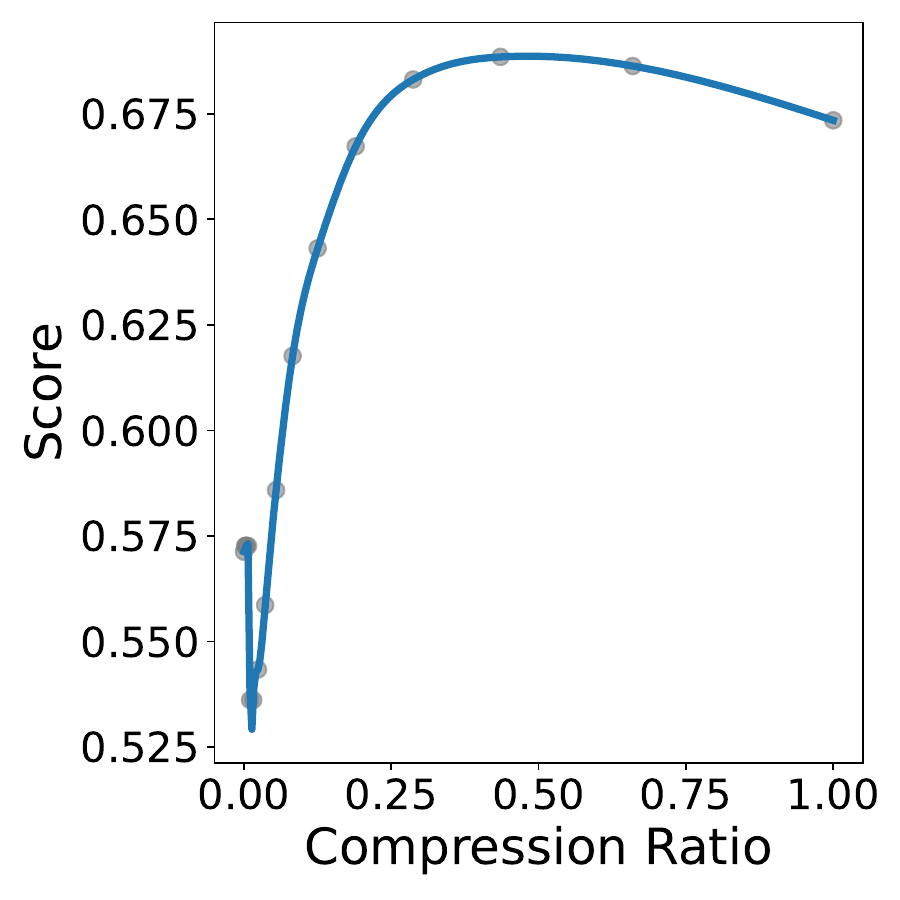}
    }
\end{minipage}

\caption{The performance-compression curve interpolated by the context-agnostic predictor. To illustrate the sample-specific variability, we randomly plot the performance-compression curves of 3 samples.  (a)--(f) present the performance-compression curves for the QA and summarization tasks, while (h) shows the unnormalized performance-compression curve (where the y-axis represents the F1 score) on TriviaQA. The unnormalized performance-compression curves for the other datasets are provided in Appendix~\ref{sec:scores}.}
\label{fig:performance_comparison}
\end{figure*}

\subsection{Performance-Compression Curve}
\label{sec:analysis_tradeoff}

To understand the dynamics of our framework and motivate the need for a context-aware approach, we analyze the performance-compression curves. Figure~\ref{fig:performance_comparison} plots these performance-compression curves for all seven benchmarks. The solid blue curve in each subplot represents the average performance retention across the entire training dataset at various compression ratios, effectively simulating a context-agnostic predictor. The dashed lines show the volatile, sample-specific performance curves for three randomly selected samples.

A primary observation from the context-agnostic curves (solid blue lines) is that for most datasets, average performance monotonically increases with the compression ratio (i.e., retaining more context). However, TriviaQA stands out as a notable exception. As shown in Figure~\ref{fig:TriviaQA_f1_s}, its average performance peaks at a compression ratio of approximately 0.435, outperforming even the uncompressed context. This suggests that the original contexts in TriviaQA contain a significant amount of low-information-density noise. A moderate compression acts as a beneficial denoising mechanism, removing distracting information and allowing the reader LLM to focus on the relevant facts, thereby improving its performance. For datasets with cleaner contexts, higher compression ratios predictably yield better average performance.

The crucial insight, however, comes from comparing the dataset-level average (solid blue line) with the individual sample curves (dashed lines). For QA tasks in particular, the performance of a single sample often exhibits a "cliff-like" behavior: performance remains perfect until a critical piece of information is removed, at which point the score drops catastrophically to zero (e.g., see samples in Figure~\ref{fig:TriviaQA} and~\ref{fig:SQuAD}). A context-agnostic predictor, which operates on the average trend, is blind to this nuance and cannot optimize for individual samples. In contrast, our context-aware predictor is designed to model these specific, non-linear curves, enabling it to apply aggressive compression where possible while remaining conservative when necessary.

Summarization tasks exhibit a different pattern. As seen in Figure~\ref{fig:gov_report} and~\ref{fig:summ_screen_fd}, the individual sample curves are generally less abrupt and more closely follow the average trend compared to QA tasks. Since summarization requires a more holistic understanding of the text, performance degrades more gracefully as tokens are removed. This smoother, more predictable relationship explains the lower PPE scores for summarization in Table~\ref{tab:summ_results_updated}, as the performance-compression curve is inherently easier to model.

\begin{table}[t]
\centering
\small
\begin{tabular}{@{}lccc@{}}
\toprule
\textbf{Component} & \textbf{Ratio} & \textbf{Latency (ms)} & \textbf{Std. Dev. (ms)} \\
\midrule
Predictor & - & 1.53 & 0.03 \\
\midrule
Compressor & 0.02 & 27.41 & 0.36 \\
& 0.10 & 27.51 & 0.69 \\
& 0.20 & 27.60 & 0.42 \\
& 0.50 & 27.68 & 0.62 \\
& 0.80 & 27.72 & 0.48 \\
& 1.00 & 27.76 & 0.64 \\
\midrule
Reader LLM & 0.02 & 30.15 & 0.80 \\
& 0.10 & 147.22 & 0.44 \\
& 0.20 & 365.38 & 1.12 \\
& 0.50 & 1561.84 & 2.10 \\
& 0.80 & 3550.31 & 5.92 \\
& 1.00 & 5282.95 & 5.76 \\
\bottomrule
\end{tabular}
\caption{Latency evaluation of the framework components. The predictor and compressor latencies represent the overhead, measured on 512-token chunks. The Reader latency demonstrates the inference speedup on a 32,768-token context compressed at different ratios.}
\label{tab:latency_evaluation}
\end{table}

\subsection{Latency Evaluation}
For our framework to be practical, its overhead must be minimal compared to the latency savings it provides. We evaluate the latency of each component, with results presented in Table~\ref{tab:latency_evaluation}. All measurements are averaged over 100 runs.

The overhead consists of two parts: the performance predictor and the compressor. The predictor adds a negligible, fixed latency of just 1.53 ms per 512-token chunk. The compressor's latency is also remarkably low and stable, averaging around 27.5 ms per chunk, irrespective of the target compression ratio. This demonstrates that determining how much to compress and performing the compression itself is extremely efficient.

The true benefit is revealed in the reader LLM's inference time. We measured the latency for processing a 32,768-token context after it was compressed to various ratios. Without compression (ratio 1.0), the reader takes 5283 ms. By applying even moderate compression, the gains are substantial. At a ratio of 0.2, latency drops to 365 ms (a 14.5x speedup), and at a ratio of 0.1, it plummets to 147 ms (a 36x speedup).

This analysis confirms that the small, fixed overhead of our framework (approximately 29 ms per 512-token chunk) is vastly outweighed by the massive, super-linear reduction in end-to-end latency.

\section{Conclusion}


We introduce Performance-oriented Context Compression (PoC), a novel paradigm that addresses the unpredictable performance degradation of existing context compression methods. Instead of a fixed compression budget, PoC allows users to specify a performance floor. A lightweight, context-aware performance predictor then automatically determines the most aggressive compression ratio for each context that meets this performance constraint.
Our experiments show this approach enhances reliability, achieves lower performance prediction error and superior overall performance compared to a context-agnostic baseline, representing a significant step toward reliable and efficient deployment of context compression for LLMs.

\section*{Limitations}

A limitation of our PoC framework is that its overall effectiveness is inherently tied to the capabilities of the underlying off-the-shelf compressor. PoC can optimize the use of a given compressor, but it cannot improve upon its fundamental compression quality. Thus, the framework's performance ceiling is determined by the compressor itself.

\bibliography{main}

@article{pandya2021question,
  title={Question answering survey: Directions, challenges, datasets, evaluation matrices},
  author={Pandya, Hariom A and Bhatt, Brijesh S},
  journal={arXiv preprint arXiv:2112.03572},
  year={2021}
}

@article{zhang2025systematic,
  title={A systematic survey of text summarization: From statistical methods to large language models},
  author={Zhang, Haopeng and Yu, Philip S and Zhang, Jiawei},
  journal={ACM Computing Surveys},
  volume={57},
  number={11},
  pages={1--41},
  year={2025},
  publisher={ACM New York, NY}
}

@article{chang2024efficient,
  title={Efficient prompting methods for large language models: A survey},
  author={Chang, Kaiyan and Xu, Songcheng and Wang, Chenglong and Luo, Yingfeng and Liu, Xiaoqian and Xiao, Tong and Zhu, Jingbo},
  journal={arXiv preprint arXiv:2404.01077},
  year={2024}
}

@inproceedings{li-etal-2025-prompt,
    title = "Prompt Compression for Large Language Models: A Survey",
    author = "Li, Zongqian  and
      Liu, Yinhong  and
      Su, Yixuan  and
      Collier, Nigel",
    editor = "Chiruzzo, Luis  and
      Ritter, Alan  and
      Wang, Lu",
    booktitle = "Proceedings of the 2025 Conference of the Nations of the Americas Chapter of the Association for Computational Linguistics: Human Language Technologies (Volume 1: Long Papers)",
    month = apr,
    year = "2025",
    address = "Albuquerque, New Mexico",
    publisher = "Association for Computational Linguistics",
    url = "https://aclanthology.org/2025.naacl-long.368/",
    doi = "10.18653/v1/2025.naacl-long.368",
    pages = "7182--7195",
    ISBN = "979-8-89176-189-6"
}

@article{lewis2020retrieval,
  title={Retrieval-augmented generation for knowledge-intensive nlp tasks},
  author={Lewis, Patrick and Perez, Ethan and Piktus, Aleksandra and Petroni, Fabio and Karpukhin, Vladimir and Goyal, Naman and K{\"u}ttler, Heinrich and Lewis, Mike and Yih, Wen-tau and Rockt{\"a}schel, Tim and others},
  journal={Advances in neural information processing systems},
  volume={33},
  pages={9459--9474},
  year={2020}
}

@inproceedings{
jiang2023llmlingua,
title={{LLML}ingua: Compressing Prompts for Accelerated Inference of Large Language Models},
author={Huiqiang Jiang and Qianhui Wu and Chin-Yew Lin and Yuqing Yang and Lili Qiu},
booktitle={The 2023 Conference on Empirical Methods in Natural Language Processing},
year={2023},
url={https://openreview.net/forum?id=ADsEdyI32n}
}

@inproceedings{pan-etal-2024-llmlingua,
    title = "{LLML}ingua-2: Data Distillation for Efficient and Faithful Task-Agnostic Prompt Compression",
    author = {Pan, Zhuoshi  and
      Wu, Qianhui  and
      Jiang, Huiqiang  and
      Xia, Menglin  and
      Luo, Xufang  and
      Zhang, Jue  and
      Lin, Qingwei  and
      R{\"u}hle, Victor  and
      Yang, Yuqing  and
      Lin, Chin-Yew  and
      Zhao, H. Vicky  and
      Qiu, Lili  and
      Zhang, Dongmei},
    editor = "Ku, Lun-Wei  and
      Martins, Andre  and
      Srikumar, Vivek",
    booktitle = "Findings of the Association for Computational Linguistics: ACL 2024",
    month = aug,
    year = "2024",
    address = "Bangkok, Thailand",
    publisher = "Association for Computational Linguistics",
    url = "https://aclanthology.org/2024.findings-acl.57/",
    doi = "10.18653/v1/2024.findings-acl.57",
    pages = "963--981"
}

@inproceedings{jiang-etal-2024-longllmlingua,
    title = "{L}ong{LLML}ingua: Accelerating and Enhancing {LLM}s in Long Context Scenarios via Prompt Compression",
    author = "Jiang, Huiqiang  and
      Wu, Qianhui  and
      Luo, Xufang  and
      Li, Dongsheng  and
      Lin, Chin-Yew  and
      Yang, Yuqing  and
      Qiu, Lili",
    editor = "Ku, Lun-Wei  and
      Martins, Andre  and
      Srikumar, Vivek",
    booktitle = "Proceedings of the 62nd Annual Meeting of the Association for Computational Linguistics (Volume 1: Long Papers)",
    month = aug,
    year = "2024",
    address = "Bangkok, Thailand",
    publisher = "Association for Computational Linguistics",
    url = "https://aclanthology.org/2024.acl-long.91/",
    doi = "10.18653/v1/2024.acl-long.91",
    pages = "1658--1677"
}

@inproceedings{
xu2024recomp,
title={{RECOMP}: Improving Retrieval-Augmented {LM}s with Context Compression and Selective Augmentation},
author={Fangyuan Xu and Weijia Shi and Eunsol Choi},
booktitle={The Twelfth International Conference on Learning Representations},
year={2024},
url={https://openreview.net/forum?id=mlJLVigNHp}
}

@inproceedings{wingate-etal-2022-prompt,
    title = "Prompt Compression and Contrastive Conditioning for Controllability and Toxicity Reduction in Language Models",
    author = "Wingate, David  and
      Shoeybi, Mohammad  and
      Sorensen, Taylor",
    editor = "Goldberg, Yoav  and
      Kozareva, Zornitsa  and
      Zhang, Yue",
    booktitle = "Findings of the Association for Computational Linguistics: EMNLP 2022",
    month = dec,
    year = "2022",
    address = "Abu Dhabi, United Arab Emirates",
    publisher = "Association for Computational Linguistics",
    url = "https://aclanthology.org/2022.findings-emnlp.412/",
    doi = "10.18653/v1/2022.findings-emnlp.412",
    pages = "5621--5634"
}

@inproceedings{chevalier-etal-2023-adapting,
    title = "Adapting Language Models to Compress Contexts",
    author = "Chevalier, Alexis  and
      Wettig, Alexander  and
      Ajith, Anirudh  and
      Chen, Danqi",
    editor = "Bouamor, Houda  and
      Pino, Juan  and
      Bali, Kalika",
    booktitle = "Proceedings of the 2023 Conference on Empirical Methods in Natural Language Processing",
    month = dec,
    year = "2023",
    address = "Singapore",
    publisher = "Association for Computational Linguistics",
    url = "https://aclanthology.org/2023.emnlp-main.232/",
    doi = "10.18653/v1/2023.emnlp-main.232",
    pages = "3829--3846"
}

@inproceedings{
ge2024incontext,
title={In-context Autoencoder for Context Compression in a Large Language Model},
author={Tao Ge and Hu Jing and Lei Wang and Xun Wang and Si-Qing Chen and Furu Wei},
booktitle={The Twelfth International Conference on Learning Representations},
year={2024},
url={https://openreview.net/forum?id=uREj4ZuGJE}
}

@article{tang2025gmsa,
  title={Gmsa: Enhancing context compression via group merging and layer semantic alignment},
  author={Tang, Jiwei and Zhang, Zhicheng and Wu, Shunlong and Ye, Jingheng and Bai, Lichen and Wang, Zitai and Lu, Tingwei and Hai, Lin and Zhao, Yiming and Zheng, Hai-Tao and others},
  journal={arXiv preprint arXiv:2505.12215},
  year={2025}
}

@inproceedings{
mu2023learning,
title={Learning to Compress Prompts with Gist Tokens},
author={Jesse Mu and Xiang Lisa Li and Noah Goodman},
booktitle={Thirty-seventh Conference on Neural Information Processing Systems},
year={2023},
url={https://openreview.net/forum?id=2DtxPCL3T5}
}

@inproceedings{
zhang2025long,
title={Long Context Compression with Activation Beacon},
author={Peitian Zhang and Zheng Liu and Shitao Xiao and Ninglu Shao and Qiwei Ye and Zhicheng Dou},
booktitle={The Thirteenth International Conference on Learning Representations},
year={2025},
url={https://openreview.net/forum?id=1eQT9OzfNQ}
}

@inproceedings{li-etal-2025-500xcompressor,
    title = "500x{C}ompressor: Generalized Prompt Compression for Large Language Models",
    author = "Li, Zongqian  and
      Su, Yixuan  and
      Collier, Nigel",
    editor = "Che, Wanxiang  and
      Nabende, Joyce  and
      Shutova, Ekaterina  and
      Pilehvar, Mohammad Taher",
    booktitle = "Proceedings of the 63rd Annual Meeting of the Association for Computational Linguistics (Volume 1: Long Papers)",
    month = jul,
    year = "2025",
    address = "Vienna, Austria",
    publisher = "Association for Computational Linguistics",
    url = "https://aclanthology.org/2025.acl-long.1219/",
    doi = "10.18653/v1/2025.acl-long.1219",
    pages = "25081--25091",
    ISBN = "979-8-89176-251-0"
}

@inproceedings{zhao-etal-2025-position,
    title = "Position {ID}s Matter: An Enhanced Position Layout for Efficient Context Compression in Large Language Models",
    author = "Zhao, Runsong  and
      Liu, Xin  and
      Liu, Xinyu  and
      Huang, Pengcheng  and
      Xiao, Chunyang  and
      Xiao, Tong  and
      Zhu, JingBo",
    editor = "Christodoulopoulos, Christos  and
      Chakraborty, Tanmoy  and
      Rose, Carolyn  and
      Peng, Violet",
    booktitle = "Findings of the Association for Computational Linguistics: EMNLP 2025",
    month = nov,
    year = "2025",
    address = "Suzhou, China",
    publisher = "Association for Computational Linguistics",
    url = "https://aclanthology.org/2025.findings-emnlp.962/",
    doi = "10.18653/v1/2025.findings-emnlp.962",
    pages = "17715--17734",
    ISBN = "979-8-89176-335-7"
}

@inproceedings{
liu2026autoencodingfree,
title={Autoencoding-Free Context Compression for {LLM}s via Contextual Semantic Anchors},
author={Xin Liu and Runsong Zhao and Pengcheng Huang and Xinyu Liu and Junyi Xiao and Chunyang Xiao and Tong Xiao and Shengxiang Gao and Zhengtao Yu and JingBo Zhu},
booktitle={The Fourteenth International Conference on Learning Representations},
year={2026},
url={https://openreview.net/forum?id=8Pi6Du0n7F}
}

@inproceedings{cao-etal-2024-retaining,
    title = "Retaining Key Information under High Compression Ratios: Query-Guided Compressor for {LLM}s",
    author = "Cao, Zhiwei  and
      Cao, Qian  and
      Lu, Yu  and
      Peng, Ningxin  and
      Huang, Luyang  and
      Cheng, Shanbo  and
      Su, Jinsong",
    editor = "Ku, Lun-Wei  and
      Martins, Andre  and
      Srikumar, Vivek",
    booktitle = "Proceedings of the 62nd Annual Meeting of the Association for Computational Linguistics (Volume 1: Long Papers)",
    month = aug,
    year = "2024",
    address = "Bangkok, Thailand",
    publisher = "Association for Computational Linguistics",
    url = "https://aclanthology.org/2024.acl-long.685/",
    doi = "10.18653/v1/2024.acl-long.685",
    pages = "12685--12695"
}

@inproceedings{guo-etal-2025-enhancing,
    title = "Enhancing {RAG} Efficiency with Adaptive Context Compression",
    author = "Guo, Shuyu  and
      Zhang, Shuo  and
      Ren, Zhaochun",
    editor = "Christodoulopoulos, Christos  and
      Chakraborty, Tanmoy  and
      Rose, Carolyn  and
      Peng, Violet",
    booktitle = "Findings of the Association for Computational Linguistics: EMNLP 2025",
    month = nov,
    year = "2025",
    address = "Suzhou, China",
    publisher = "Association for Computational Linguistics",
    url = "https://aclanthology.org/2025.findings-emnlp.1307/",
    doi = "10.18653/v1/2025.findings-emnlp.1307",
    pages = "24061--24076",
    ISBN = "979-8-89176-335-7"
}

@article{zhang2024adacomp,
  title={Adacomp: Extractive context compression with adaptive predictor for retrieval-augmented large language models},
  author={Zhang, Qianchi and Zhang, Hainan and Pang, Liang and Zheng, Hongwei and Zheng, Zhiming},
  journal={arXiv preprint arXiv:2409.01579},
  year={2024}
}

@inproceedings{luo-etal-2025-attncomp,
    title = "{A}ttn{C}omp: Attention-Guided Adaptive Context Compression for Retrieval-Augmented Generation",
    author = "Luo, Lvzhou  and
      Cao, Yixuan  and
      Luo, Ping",
    editor = "Christodoulopoulos, Christos  and
      Chakraborty, Tanmoy  and
      Rose, Carolyn  and
      Peng, Violet",
    booktitle = "Findings of the Association for Computational Linguistics: EMNLP 2025",
    month = nov,
    year = "2025",
    address = "Suzhou, China",
    publisher = "Association for Computational Linguistics",
    url = "https://aclanthology.org/2025.findings-emnlp.449/",
    doi = "10.18653/v1/2025.findings-emnlp.449",
    pages = "8456--8472",
    ISBN = "979-8-89176-335-7"
}

@inproceedings{chen-etal-2025-dast,
    title = "{DAST}: Context-Aware Compression in {LLM}s via Dynamic Allocation of Soft Tokens",
    author = "Chen, Shaoshen  and
      Li, Yangning  and
      Xu, Zishan  and
      Zeng, Yongqin  and
      Wu, Shunlong  and
      Hu, Xinshuo  and
      Shan, Zifei  and
      Su, Xin  and
      Tang, Jiwei  and
      Li, Yinghui  and
      Zheng, Hai-Tao",
    editor = "Che, Wanxiang  and
      Nabende, Joyce  and
      Shutova, Ekaterina  and
      Pilehvar, Mohammad Taher",
    booktitle = "Findings of the Association for Computational Linguistics: ACL 2025",
    month = jul,
    year = "2025",
    address = "Vienna, Austria",
    publisher = "Association for Computational Linguistics",
    url = "https://aclanthology.org/2025.findings-acl.1055/",
    doi = "10.18653/v1/2025.findings-acl.1055",
    pages = "20544--20552",
    ISBN = "979-8-89176-256-5"
}

@article{tang2026read,
  title={Read As Human: Compressing Context via Parallelizable Close Reading and Skimming},
  author={Tang, Jiwei and Liu, Shilei and Zhang, Zhicheng and Lv, Qingsong and Zhao, Runsong and Lu, Tingwei and Liu, Langming and Chen, Haibin and Yuan, Yujin and Zheng, Hai-Tao and others},
  journal={arXiv preprint arXiv:2602.01840},
  year={2026}
}

@inproceedings{rajpurkar-etal-2016-squad,
    title = "{SQ}u{AD}: 100,000+ Questions for Machine Comprehension of Text",
    author = "Rajpurkar, Pranav  and
      Zhang, Jian  and
      Lopyrev, Konstantin  and
      Liang, Percy",
    editor = "Su, Jian  and
      Duh, Kevin  and
      Carreras, Xavier",
    booktitle = "Proceedings of the 2016 Conference on Empirical Methods in Natural Language Processing",
    month = nov,
    year = "2016",
    address = "Austin, Texas",
    publisher = "Association for Computational Linguistics",
    url = "https://aclanthology.org/D16-1264/",
    doi = "10.18653/v1/D16-1264",
    pages = "2383--2392"
}

@article{dunn2017searchqa,
  title={Searchqa: A new q\&a dataset augmented with context from a search engine},
  author={Dunn, Matthew and Sagun, Levent and Higgins, Mike and Guney, V Ugur and Cirik, Volkan and Cho, Kyunghyun},
  journal={arXiv preprint arXiv:1704.05179},
  year={2017}
}

@inproceedings{joshi-etal-2017-triviaqa,
    title = "{T}rivia{QA}: A Large Scale Distantly Supervised Challenge Dataset for Reading Comprehension",
    author = "Joshi, Mandar  and
      Choi, Eunsol  and
      Weld, Daniel  and
      Zettlemoyer, Luke",
    editor = "Barzilay, Regina  and
      Kan, Min-Yen",
    booktitle = "Proceedings of the 55th Annual Meeting of the Association for Computational Linguistics (Volume 1: Long Papers)",
    month = jul,
    year = "2017",
    address = "Vancouver, Canada",
    publisher = "Association for Computational Linguistics",
    url = "https://aclanthology.org/P17-1147/",
    doi = "10.18653/v1/P17-1147",
    pages = "1601--1611"
}

@article{kwiatkowski-etal-2019-natural,
    title = "Natural Questions: A Benchmark for Question Answering Research",
    author = "Kwiatkowski, Tom  and
      Palomaki, Jennimaria  and
      Redfield, Olivia  and
      Collins, Michael  and
      Parikh, Ankur  and
      Alberti, Chris  and
      Epstein, Danielle  and
      Polosukhin, Illia  and
      Devlin, Jacob  and
      Lee, Kenton  and
      Toutanova, Kristina  and
      Jones, Llion  and
      Kelcey, Matthew  and
      Chang, Ming-Wei  and
      Dai, Andrew M.  and
      Uszkoreit, Jakob  and
      Le, Quoc  and
      Petrov, Slav",
    editor = "Lee, Lillian  and
      Johnson, Mark  and
      Roark, Brian  and
      Nenkova, Ani",
    journal = "Transactions of the Association for Computational Linguistics",
    volume = "7",
    year = "2019",
    address = "Cambridge, MA",
    publisher = "MIT Press",
    url = "https://aclanthology.org/Q19-1026/",
    doi = "10.1162/tacl_a_00276",
    pages = "452--466"
}

@inproceedings{yang-etal-2018-hotpotqa,
    title = "{H}otpot{QA}: A Dataset for Diverse, Explainable Multi-hop Question Answering",
    author = "Yang, Zhilin  and
      Qi, Peng  and
      Zhang, Saizheng  and
      Bengio, Yoshua  and
      Cohen, William  and
      Salakhutdinov, Ruslan  and
      Manning, Christopher D.",
    editor = "Riloff, Ellen  and
      Chiang, David  and
      Hockenmaier, Julia  and
      Tsujii, Jun{'}ichi",
    booktitle = "Proceedings of the 2018 Conference on Empirical Methods in Natural Language Processing",
    month = oct # "-" # nov,
    year = "2018",
    address = "Brussels, Belgium",
    publisher = "Association for Computational Linguistics",
    url = "https://aclanthology.org/D18-1259/",
    doi = "10.18653/v1/D18-1259",
    pages = "2369--2380"
}

@inproceedings{huang-etal-2021-efficient,
    title = "Efficient Attentions for Long Document Summarization",
    author = "Huang, Luyang  and
      Cao, Shuyang  and
      Parulian, Nikolaus  and
      Ji, Heng  and
      Wang, Lu",
    editor = "Toutanova, Kristina  and
      Rumshisky, Anna  and
      Zettlemoyer, Luke  and
      Hakkani-Tur, Dilek  and
      Beltagy, Iz  and
      Bethard, Steven  and
      Cotterell, Ryan  and
      Chakraborty, Tanmoy  and
      Zhou, Yichao",
    booktitle = "Proceedings of the 2021 Conference of the North American Chapter of the Association for Computational Linguistics: Human Language Technologies",
    month = jun,
    year = "2021",
    address = "Online",
    publisher = "Association for Computational Linguistics",
    url = "https://aclanthology.org/2021.naacl-main.112/",
    doi = "10.18653/v1/2021.naacl-main.112",
    pages = "1419--1436"
}

@inproceedings{lin-2004-rouge,
    title = "{ROUGE}: A Package for Automatic Evaluation of Summaries",
    author = "Lin, Chin-Yew",
    booktitle = "Text Summarization Branches Out",
    month = jul,
    year = "2004",
    address = "Barcelona, Spain",
    publisher = "Association for Computational Linguistics",
    url = "https://aclanthology.org/W04-1013/",
    pages = "74--81"
}

@inproceedings{chen-etal-2022-summscreen,
    title = "{S}umm{S}creen: A Dataset for Abstractive Screenplay Summarization",
    author = "Chen, Mingda  and
      Chu, Zewei  and
      Wiseman, Sam  and
      Gimpel, Kevin",
    editor = "Muresan, Smaranda  and
      Nakov, Preslav  and
      Villavicencio, Aline",
    booktitle = "Proceedings of the 60th Annual Meeting of the Association for Computational Linguistics (Volume 1: Long Papers)",
    month = may,
    year = "2022",
    address = "Dublin, Ireland",
    publisher = "Association for Computational Linguistics",
    url = "https://aclanthology.org/2022.acl-long.589/",
    doi = "10.18653/v1/2022.acl-long.589",
    pages = "8602--8615"
}

@article{grattafiori2024llama,
  title={The llama 3 herd of models},
  author={Grattafiori, Aaron and Dubey, Abhimanyu and Jauhri, Abhinav and Pandey, Abhinav and Kadian, Abhishek and Al-Dahle, Ahmad and Letman, Aiesha and Mathur, Akhil and Schelten, Alan and Vaughan, Alex and others},
  journal={arXiv preprint arXiv:2407.21783},
  year={2024}
}

@inproceedings{
loshchilov2018decoupled,
title={Decoupled Weight Decay Regularization},
author={Ilya Loshchilov and Frank Hutter},
booktitle={International Conference on Learning Representations},
year={2019},
url={https://openreview.net/forum?id=Bkg6RiCqY7},
}

@misc{tang2025perceptioncompressortrainingfreeprompt,
      title={Perception Compressor: A Training-Free Prompt Compression Framework in Long Context Scenarios}, 
      author={Jiwei Tang and Jin Xu and Tingwei Lu and Zhicheng Zhang and Yiming Zhao and Lin Hai and Hai-Tao Zheng},
      year={2025},
      eprint={2409.19272},
      archivePrefix={arXiv},
      primaryClass={cs.CL},
      url={https://arxiv.org/abs/2409.19272}, 
}

@misc{tang2026comicoarsetofinecontextcompression,
      title={COMI: Coarse-to-fine Context Compression via Marginal Information Gain}, 
      author={Jiwei Tang and Shilei Liu and Zhicheng Zhang and Yujin Yuan and Libin Zheng and Wenbo Su and Bo Zheng},
      year={2026},
      eprint={2602.01719},
      archivePrefix={arXiv},
      primaryClass={cs.CL},
      url={https://arxiv.org/abs/2602.01719}, 
}

@misc{lv2026datadistributionmattersdatacentric,
      title={Data Distribution Matters: A Data-Centric Perspective on Context Compression for Large Language Model}, 
      author={Kangtao Lv and Jiwei Tang and Langming Liu and Haibin Chen and Weidong Zhang and Shilei Liu and Yongwei Wang and Yujin Yuan and Wenbo Su and Bo Zheng},
      year={2026},
      eprint={2602.01778},
      archivePrefix={arXiv},
      primaryClass={cs.CL},
      url={https://arxiv.org/abs/2602.01778}, 
}

@inproceedings{liu-etal-2024-forgetting,
    title = "Forgetting Curve: A Reliable Method for Evaluating Memorization Capability for Long-Context Models",
    author = "Liu, Xinyu  and
      Zhao, Runsong  and
      Huang, Pengcheng  and
      Xiao, Chunyang  and
      Li, Bei  and
      Wang, Jingang  and
      Xiao, Tong  and
      Zhu, JingBo",
    editor = "Al-Onaizan, Yaser  and
      Bansal, Mohit  and
      Chen, Yun-Nung",
    booktitle = "Proceedings of the 2024 Conference on Empirical Methods in Natural Language Processing",
    month = nov,
    year = "2024",
    address = "Miami, Florida, USA",
    publisher = "Association for Computational Linguistics",
    url = "https://aclanthology.org/2024.emnlp-main.269/",
    doi = "10.18653/v1/2024.emnlp-main.269",
    pages = "4667--4682"
}

@article{zhao2026comet,
  title={CoMeT: Collaborative Memory Transformer for Efficient Long Context Modeling},
  author={Zhao, Runsong and Liu, Shilei and Tang, Jiwei and Liu, Langming and Chen, Haibin and Zhang, Weidong and Yuan, Yujin and Xiao, Tong and Zhu, Jingbo and Su, Wenbo and others},
  journal={arXiv preprint arXiv:2602.01766},
  year={2026}
}

\appendix

\begin{figure*}[t]
\centering
\captionsetup[sub]{font=small, labelfont=small}

 \begin{minipage}{0.25\textwidth}
        \centering
        \subcaptionbox{SearchQA\label{fig:SearchQA_f1}}{
         \includegraphics[width=\linewidth]{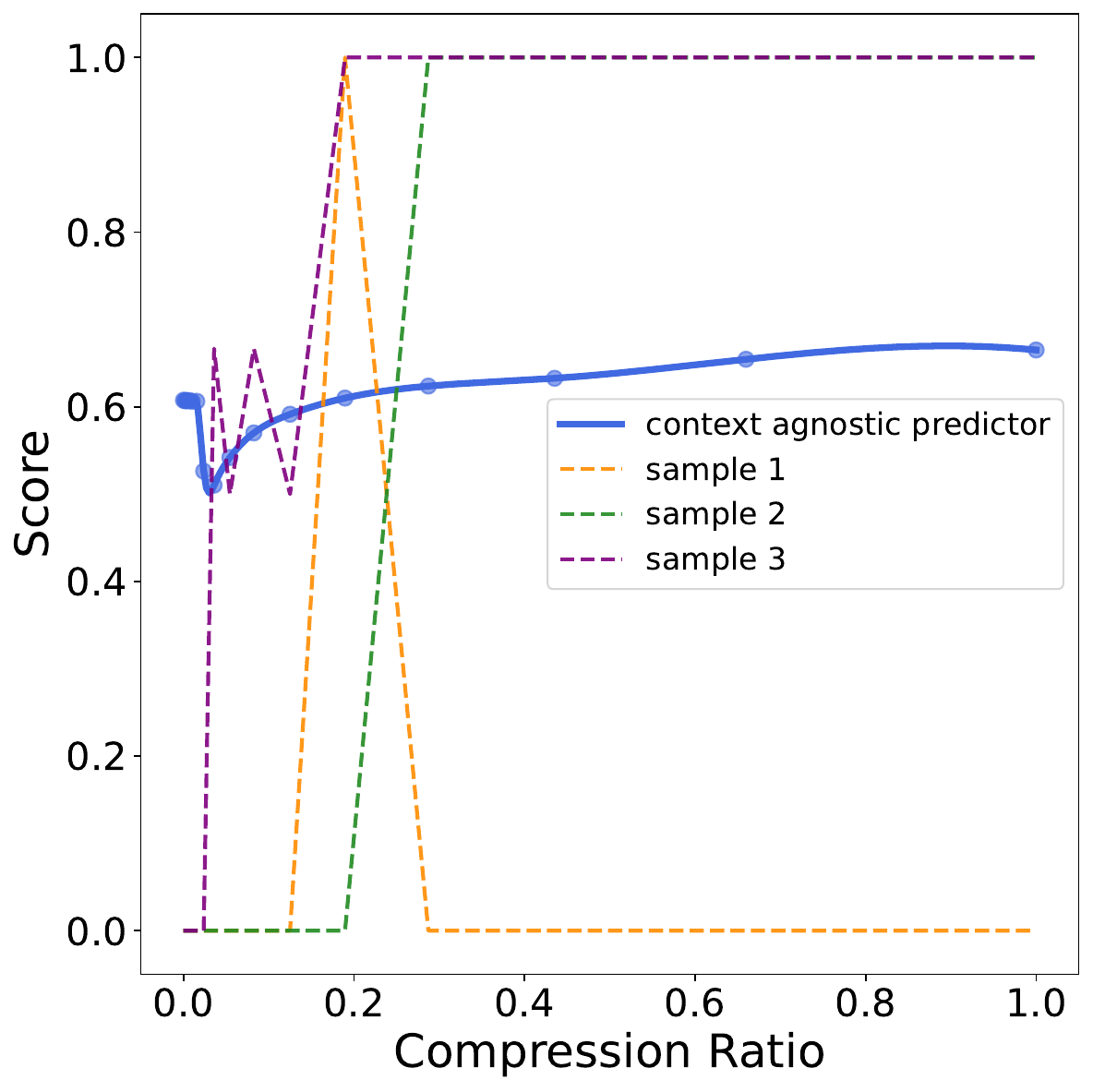}
        }
 \end{minipage}%
 \begin{minipage}{0.25\textwidth}
        \centering
        \subcaptionbox{TriviaQA\label{fig:TriviaQA_f1}}{
         \includegraphics[width=\linewidth]{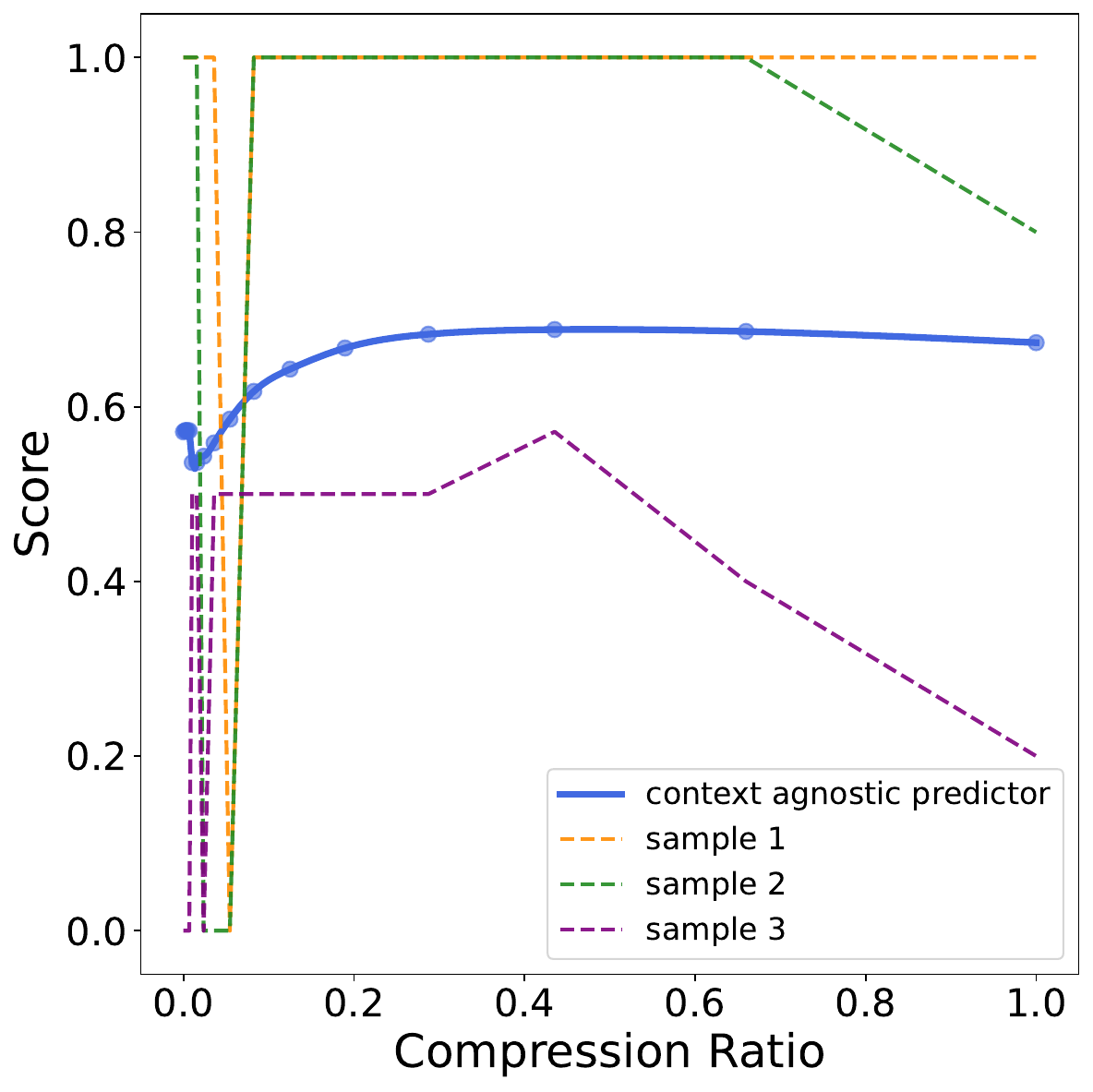}
        }
 \end{minipage}%
 \begin{minipage}{0.25\textwidth}
    \centering
    \subcaptionbox{Natural Questions\label{fig:NQ_f1}}{
    \includegraphics[width=\linewidth]{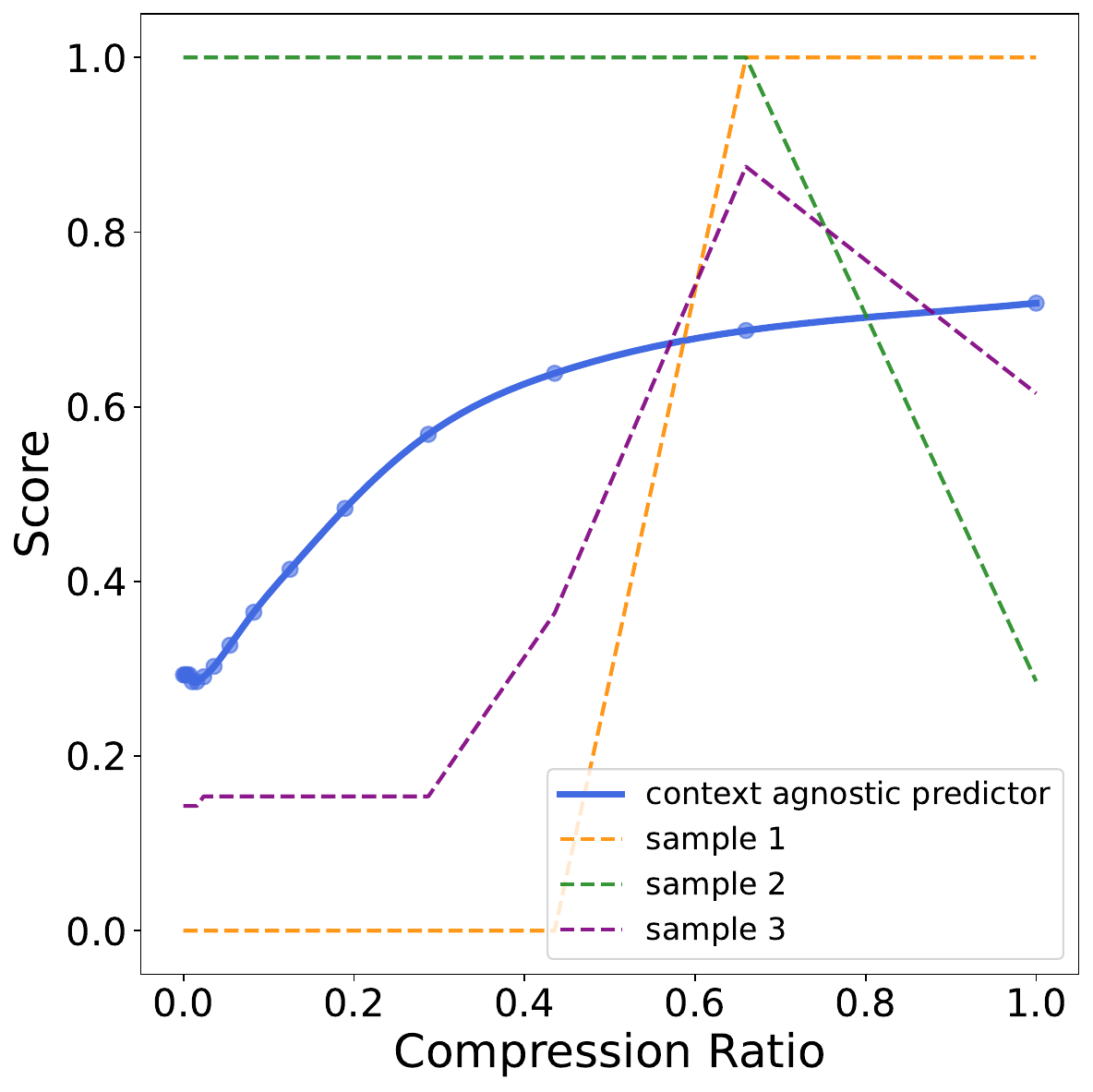}
    }
\end{minipage}%
\begin{minipage}{0.25\textwidth}
    \centering
    \subcaptionbox{HotpotQA\label{fig:HotpotQA_f1}}{
    \includegraphics[width=\linewidth]{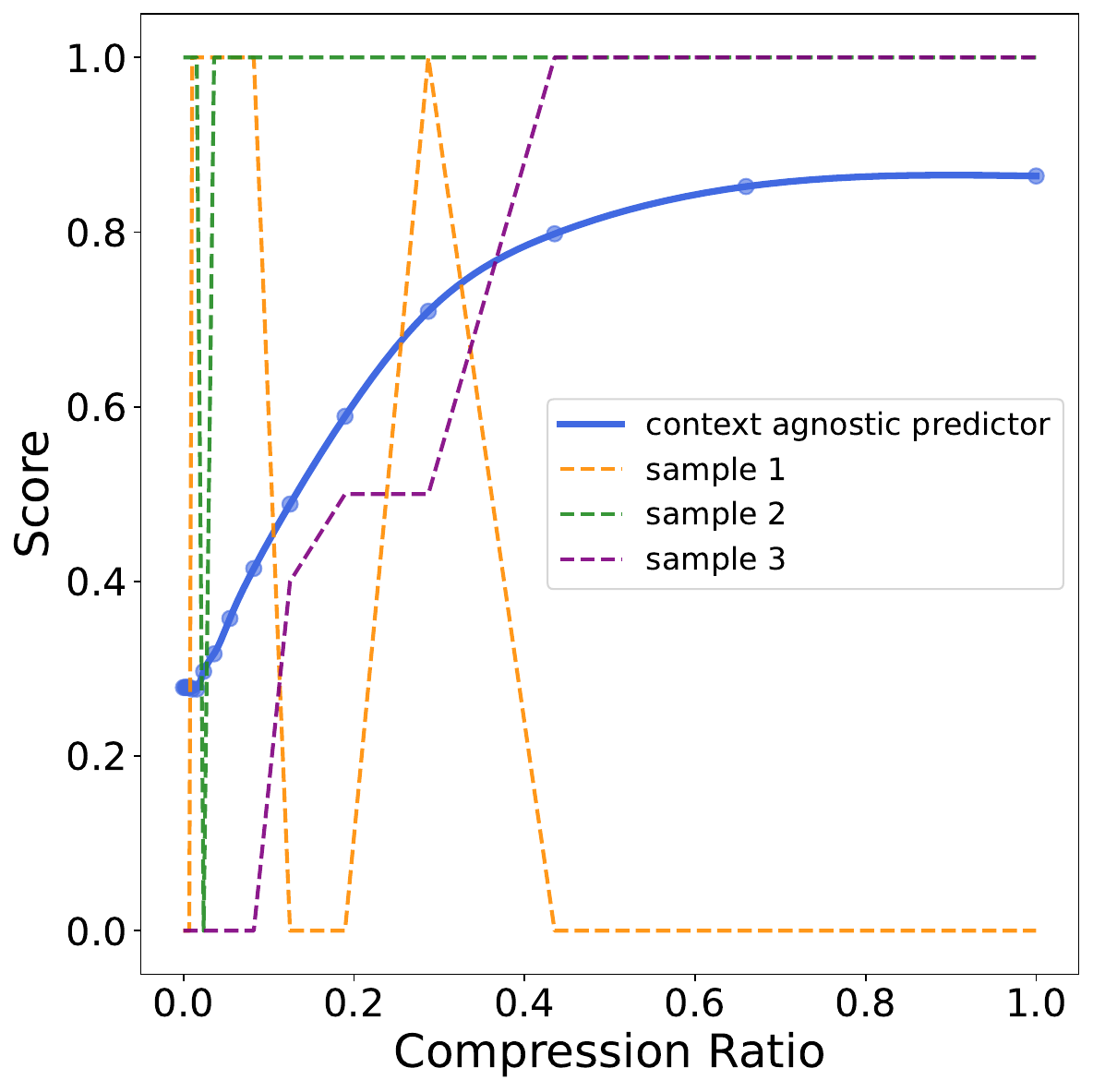}
    }
\end{minipage}

\vspace{5mm}

 \begin{minipage}{0.25\textwidth}
        \centering
        \subcaptionbox{SQuAD\label{fig:SQuAD_f1}}{
         \includegraphics[width=\linewidth]{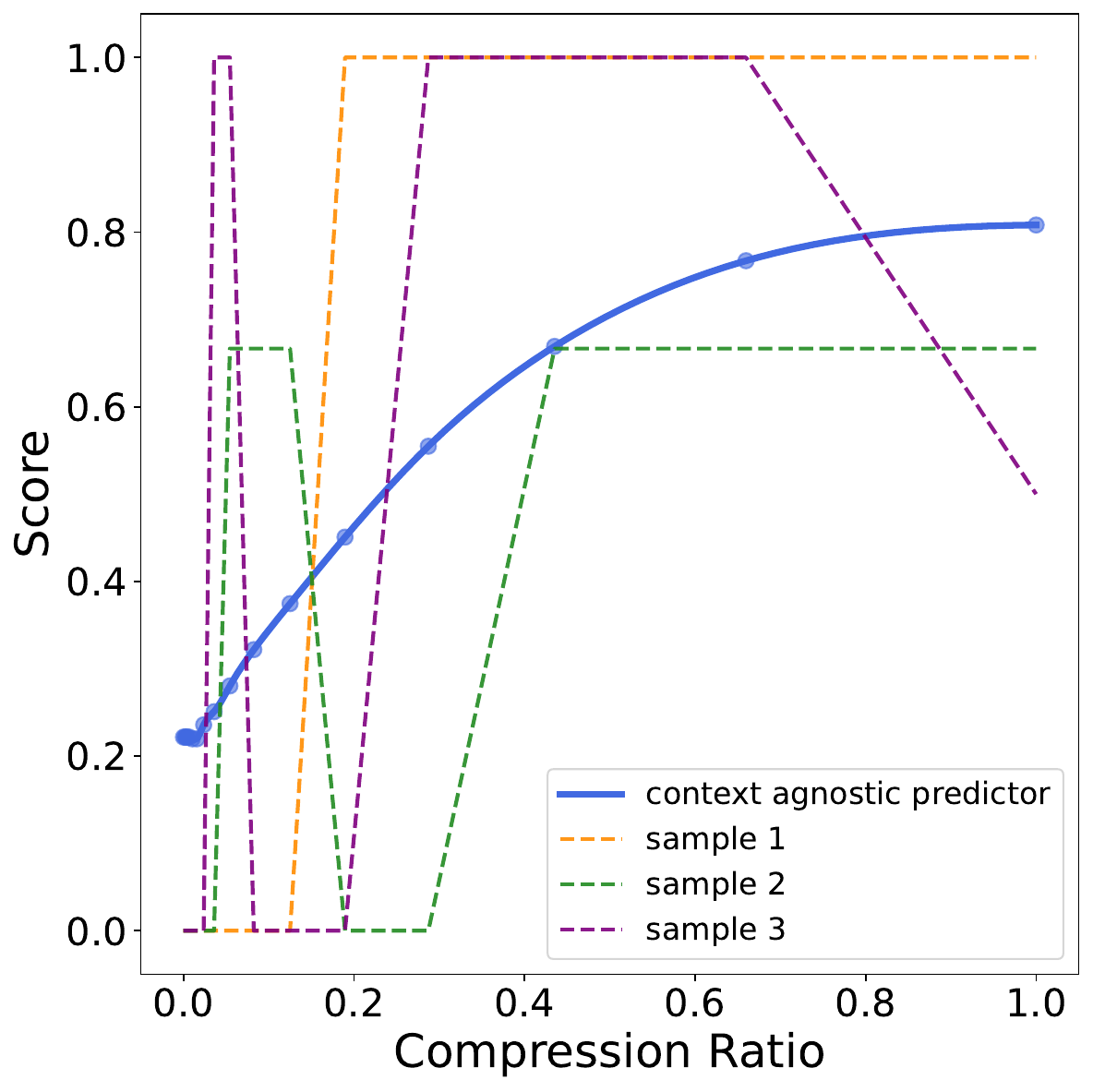}
        }
 \end{minipage}%
 \begin{minipage}{0.25\textwidth}
        \centering
        \subcaptionbox{GovReport\label{fig:gov_report_rouge}}{
         \includegraphics[width=\linewidth]{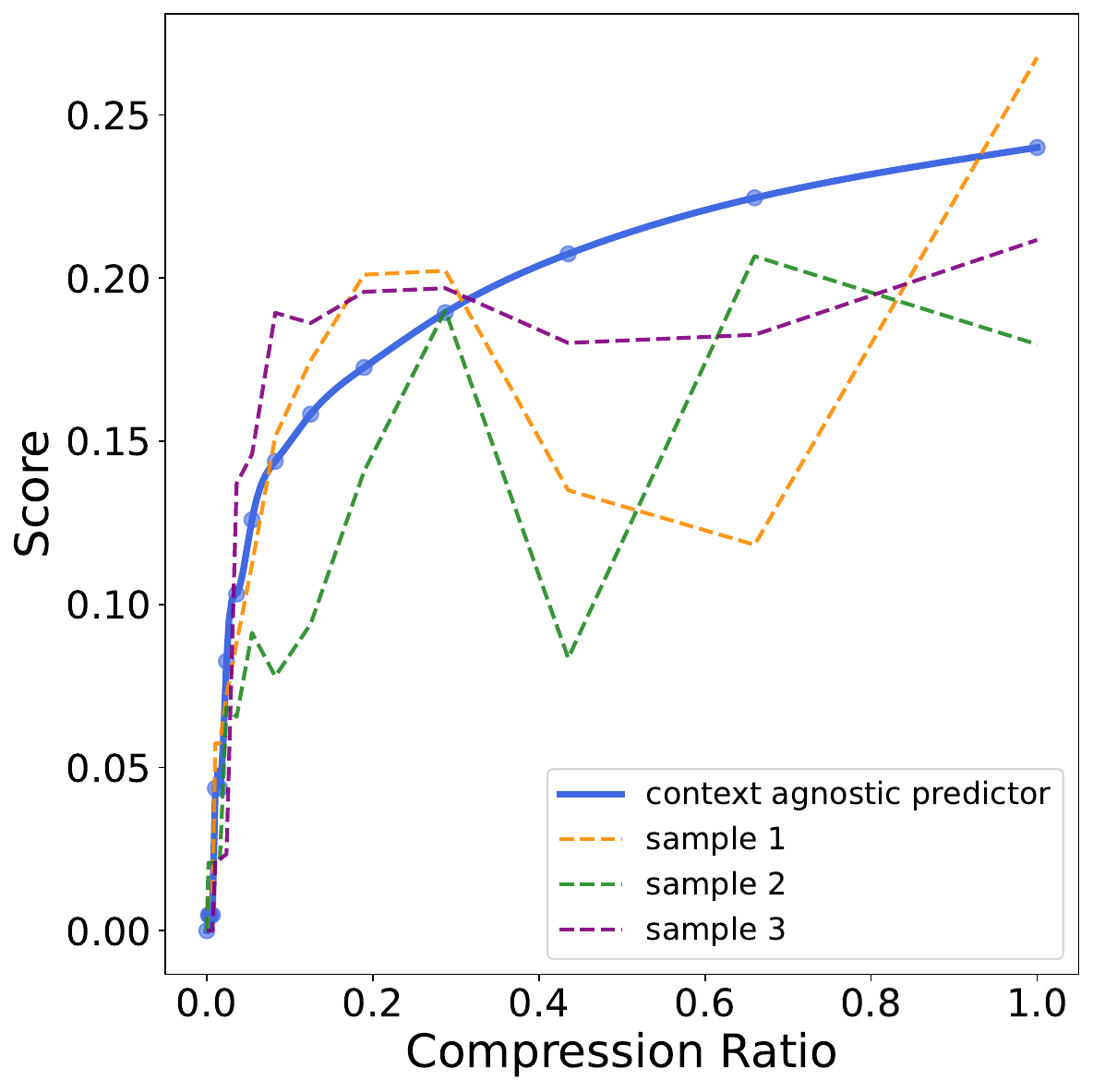}
        }
 \end{minipage}%
 \begin{minipage}{0.25\textwidth}
    \centering
    \subcaptionbox{SummScreenFD\label{fig:summ_screen_fd_rouge}}{
    \includegraphics[width=\linewidth]{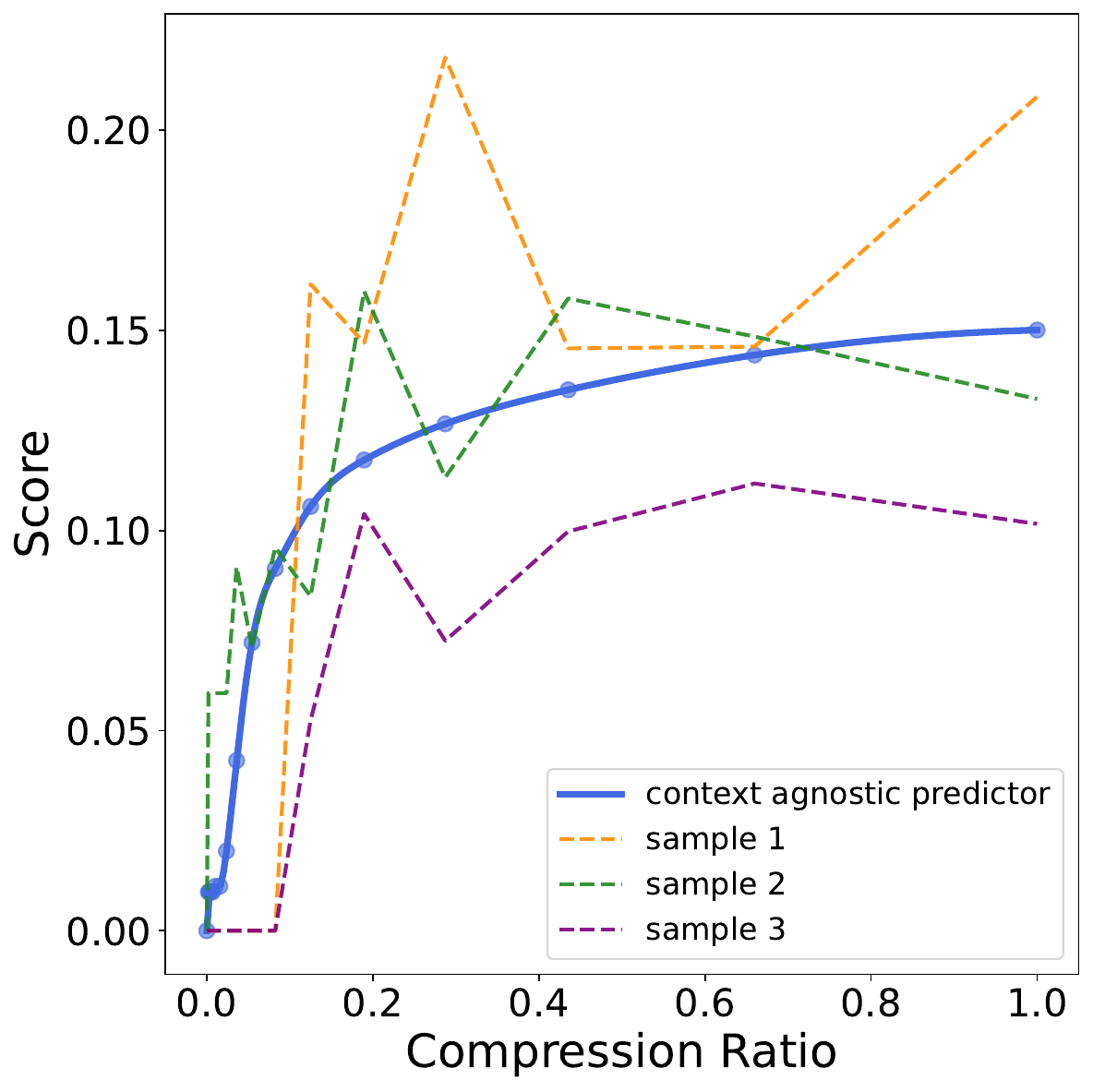}
    }
\end{minipage}%
\begin{minipage}{0.25\textwidth}
    \centering
    \subcaptionbox{TriviaQA\label{fig:TriviaQA_f1_no}}{
    \includegraphics[width=\linewidth]{plots/llmlingua_TriviaQA_minimal.pdf}
    }
\end{minipage}

\caption{The unnormalized performance-compression curves interpolated by the context-agnostic predictor. In subfigures (a)--(e), the y-axis represents the F1 score, while in (f) and (g), the y-axis represents the geometric mean of ROUGE-1, ROUGE-2, and ROUGE-L.
}
\label{fig:unnorm_performance_comparison}
\end{figure*}

\section{Unnormalized Performance Curves at Different Compression Ratios}
\label{sec:scores}
Figure~\ref{fig:unnorm_performance_comparison} presents the unnormalized scores across the 7 evaluated datasets at varying compression ratios, corresponding to Figure~\ref{fig:performance_comparison}.

\end{document}